\documentclass{article}
\usepackage{lineno}
\usepackage{wrapfig}

\usepackage{graphicx}
\usepackage{floatrow}
\usepackage{sidecap}
\usepackage{caption}
\usepackage{calc} 

\usepackage[numbers]{natbib}

\usepackage[preprint]{neurips_2022}
\usepackage{xcolor}         
\usepackage[utf8]{inputenc} 
\usepackage[T1]{fontenc}    
\usepackage{hyperref}       
\usepackage{url}            
\usepackage{booktabs}       
\usepackage{tablefootnote}
\usepackage{amsfonts}       
\usepackage{nicefrac}       
\usepackage{microtype}      

\definecolor{myblue}{rgb}{0.255, 0.412, 0.882} 

\definecolor{green1}{RGB}{10,158,10}
\definecolor{blue1}{RGB}{17,85,204}
\definecolor{red1}{RGB}{204,0,0}
\definecolor{mygray}{gray}{0.85}

\definecolor{myblue2}{RGB}{187, 213, 232} 

\usepackage{hyperref}       
\usepackage{url}            
\usepackage{booktabs}       
\usepackage{amsfonts}       
\usepackage{nicefrac}       
\usepackage{microtype}      
\usepackage{lipsum}
\usepackage{wrapfig}
\usepackage{siunitx}
\usepackage{sidecap}

\usepackage{amsmath,amsfonts,amssymb,amsthm}
\usepackage{mathtools}
\usepackage{multirow}

\usepackage{bbm}
\usepackage{colortbl}
\usepackage{booktabs}
\usepackage{multirow}
\usepackage{siunitx}
\usepackage{array}

\usepackage{subcaption}

\usepackage{enumitem}

\newcolumntype{P}[1]{>{\hspace{1ex}}p{#1}<{\hspace{1ex}}}

\newtheorem{theorem}{Theorem}

\newtheorem{prop}[theorem]{Proposition}

\newtheorem{definition}[theorem]{Definition}

\title{Feature Learning in Image Hierarchies using Functional Maximal Correlation}

\author{%
  Bo Hu, \; Yuheng Bu,\; Jos{\'e}~C.~Pr{\'\i}ncipe\\
  Department of Electrical and Computer Engineering\\
  University of Florida\\
  \texttt{\{hubo, buyuheng\}@ufl.edu\; principe@cnel.ufl.edu }\\}

\begin{document}

\maketitle

\begin{abstract}


This paper proposes the Hierarchical Functional Maximal Correlation Algorithm (HFMCA), a hierarchical methodology that characterizes dependencies across two hierarchical levels in multiview systems. By framing view similarities as dependencies and ensuring contrastivity by imposing orthonormality, HFMCA achieves faster convergence and increased stability in self-supervised learning. HFMCA defines and measures dependencies within image hierarchies, from pixels and patches to full images. We find that the network topology for approximating orthonormal basis functions aligns with a vanilla CNN, enabling the decomposition of density ratios between neighboring layers of feature maps. This approach provides powerful interpretability, revealing the resemblance between supervision and self-supervision through the lens of internal representations.

\end{abstract}

\section{Introduction}

Measures of statistical dependence have been instrumental in learning codes, features, and representations that maximize information transfer~(\cite{renyi1959measures, renyi_on_measure_of_information, kraskov2004estimating, cover2006elements}). These concepts have sparked a multitude of learning principles and algorithms across various domains, including communication, neuroscience, and machine learning~(\cite{linsker1988self, wang2016deep, hjelm2019learning, saxe2019information, alemideep}). Enhancing the versatility and interpretability of multivariate features is of utmost importance in numerous machine learning tasks, including one-shot learning, transfer learning, and especially self-supervised learning~\cite{oord2018representation, chen2020simple, bardes2022vicregl}. A persistent challenge that self-supervised learning grapples with is feature collapse, a phenomenon where the model fails to learn a diverse, meaningful representation and instead, produces similar or identical output features for all input images. To mitigate this issue, various strategies have been proposed, such as the momentum-based approach~\cite{grill2020bootstrap, he2020momentum}, output decorrelation techniques~\cite{hua2021feature}, and employing information maximization criteria~\cite{ozsoy2022self}. Despite these advances, preventing feature collapse in self-supervised learning remains a significant challenge.

A recent method, Functional Maximal Correlation (FMC), introduces a novel statistical dependence measure derived from the orthonormal decomposition of the density ratio~\cite{huang2018gaussian, huang2019universal, hu2022cross}. Within this decomposition, the spectrum represents statistical dependence, while the basis functions represent the features. Together, they form the decomposition and construct the projection space via linear combinations of the basis functions. This newly proposed Functional Maximal Correlation Algorithm (FMCA), as detailed in~\cite{hu2022cross}, leverages the power of neural networks to learn this decomposition directly from empirical data. FMCA unifies the measurement of statistical dependence and feature learning by decomposing density ratios, enabling the learning of multivariate features that are theoretically orthonormal. 

We pose the question: where does statistical dependence emerge in a multiview system of self-supervised learning? Given that various diverse augmentations are generated to represent the same source object, our proposition is that the dependence effectively lies between each individual view and the collection of multiple views across two different scales. This type of dependency, which bears a hierarchical structure, requires the introduction of an adaptation to FMCA. Thus we propose the Hierarchical Functional Maximal Correlation Algorithm (HFMCA). 

The second part of the paper extends HFMCA to multiple hierarchical levels. We observe that external supervision can be naturally included as an extra source of dependence. 
We establish that the network topology required for approximating the basis functions coincides with a vanilla Convolutional Neural Network (CNN). By employing HFMCA to image hierarchies via CNN training, we manage to decompose density ratios between neighboring scales of CNN feature maps, effectively exposing internal dependence relationships. This leads to powerful interpretability, enabling us to identify the close resemblance between supervision and self-supervision through the lens of internal representations.

\textbf{Related work on self-supervised learning:} Self-supervised learning employs multivariate nonlinear mappers and similarity measures, often with moving-average moments~\cite{he2020momentum} and local structures~\cite{chen2020simple, bardes2022vicregl}. Utilizing contrastive learning~\cite{oord2018representation}, it maximizes similarity between image augmentations while contrasting different images. Common similarity measures including cosine similarity~\cite{chen2020simple, grill2020bootstrap, he2020momentum, chen2021exploring}, Euclidean distance, and matrix norms~\cite{zbontar2021barlow, wang2021vicreg, shwartz2023information} leverage the linear span of nonlinear features. 
Our approach is also fundamentally different from the existing information maximization approaches, such as~\cite{hua2021feature, ozsoy2022self}, which is still limited to two-view contrastivity. HFMCA aligns more with the recent studies that investigated multiview contrastivity~\cite{pototzky2022fastsiam, tong2023emp}.

\section{Preliminary: The Functional Maximal Correlation Algorithm}



\noindent \textbf{Spectrum, basis functions, and density ratios.} A unique characteristic of Functional Maximal Correlation (FMC)  is its direct application of spectral decomposition to the density ratio~\cite{hu2022cross}. Given any two random processes $\mathbf{X}$ and $\mathbf{Y}$, with a joint distribution $p(X, Y)$ and the marginal product $p(X)p(Y)$, FMC is defined via an orthonormal decomposition:
\begin{equation}
\resizebox{.9\linewidth}{!}{
$\begin{gathered}
\rho:= \frac{p(X, Y)}{p(X)p(Y)} = \sum_{k=1}^\infty \sqrt{\sigma_k} {\phi_k} (X) {\psi_k} (Y),\; \mathbb{E}_\mathbf{\mathbf{X}}[{\phi_k}(\mathbf{X}){\phi_{k'}}(\mathbf{X})] = \mathbb{E}_\mathbf{\mathbf{Y}}[{\psi_k}(\mathbf{Y}){\psi_{k'}}(\mathbf{Y})] = \begin{cases} 1, \;k=k'& \\ 0, \;k\neq k' \end{cases}\hspace{-10pt}.
\end{gathered}$}
\label{theory_decompose}
\end{equation}
Each decomposition component has a unique role: the spectrum measures multivariate dependence (termed FMC), the bases serve as feature projectors, and the kernel-associated density ratio provides a metric distance. The task of modeling dependence thus becomes modeling this projection space, leading to the proposal of the Functional Maximal Correlation Algorithm (FMCA).



\noindent \textbf{Neural networks implementation.} When dealing with empirical data and lacking the knowledge of $pdf$, spectral decomposition can be achieved through optimization. The empirical studies suggest a log-determinant-based cost function, optimized via paired neural networks, offers superior stability. Using two neural networks, $\mathbf{f}_\theta:\mathcal{X} \rightarrow \mathbb{R}^K$ and $\mathbf{g}_\omega:\mathcal{Y} \rightarrow \mathbb{R}^K$ that map realizations of $\mathbf{X}$ and $\mathbf{Y}$ respectively, each to a $K$-dimensional output space, we compute the autocorrelation (ACFs) and crosscorrelation functions (CCFs) defined as follows:
\begin{equation}
\resizebox{1\linewidth}{!}{
$\begin{gathered}
\mathbf{R}_F = \mathbb{E}_\mathbf{X}[\mathbf{f}_\theta(\mathbf{X})  \mathbf{f}_\theta^\intercal(\mathbf{X}) ],\; \mathbf{R}_G = \mathbb{E}_\mathbf{Y}[\mathbf{g}_\omega(\mathbf{Y})  \mathbf{g}_\omega^\intercal(\mathbf{Y}) ],\;
\mathbf{P}_{FG} = \mathbb{E}_{\mathbf{X}, \mathbf{Y}}[\mathbf{f}_\theta(\mathbf{X})  \mathbf{g}_\omega^\intercal(\mathbf{Y}) ],\; \mathbf{R}_{FG} = \begin{bmatrix}\mathbf{R}_F & \mathbf{P}_{FG} \\
\mathbf{P}^\intercal_{FG} & \mathbf{R}_G
\end{bmatrix}.
\end{gathered}$}
\label{fmca_def}
\end{equation}
FMCA defines an optimization problem that minimizes the log-determinant of the marginal ACFs $\mathbf{R}_F$ and $\mathbf{R}_G$ for output orthonormality, while maximizing the log-determinant of the joint ACF $\mathbf{R}_{FG}$ to parallelize two projection spaces. The problem is formulated as follows:
\begin{equation}
\begin{gathered}
\min_{\theta, \omega} \;r(\mathbf{f}_\theta, \mathbf{g}_\omega)  := \log\det \mathbf{R}_{FG}-\log \det \mathbf{R}_{F} - \log\det \mathbf{R}_{G}.
\end{gathered}
\label{fmca_maximal}
\end{equation}  
Upon reaching optimality, normalization schemes are employed to network outputs. Theoretically, the objective function effectively captures the leading eigenvalues of the spectrum, while the neural networks, viewed as multivariate function approximators, approximate leading basis functions.

\noindent \textbf{Linking dependence measurement and feature learning.} Applying FMCA for feature learning is direct: formulate the joint density and marginal products, initiate nonlinear mappers, and minimize costs. This yields a multivariate dependence measure and a theoretically-grounded feature projector that together decompose the density ratio. These features naturally display orthonormality, ensuring diversity, which is vital for many learning tasks. This paper spotlights the potential of this property in learning settings that exhibit \textit{hierarchical structures}. We demonstrate that learning dependence structures from hierarchies is crucial and can be proficiently accomplished with the FMCA.


\noindent \textbf{Costs, spectrum, and dependencies.} The spectrum's eigenvalues range from $0$ to $1$. The optimal cost approximates their aggregation $r^* = \sum_{k=1}^K \log (1-\sigma_k)$. Dependence can be evaluated using both the spectrum and cost, where a lower cost and higher eigenvalues indicate stronger dependence.

\vspace{-1pt}

\section{FMCA for Multiview Systems: Dependencies \& Orthonormalities}
\vspace{-1pt}




We investigate the feasibility of framing the task of self-supervised learning as a dependence measurement problem: regardless of the rotation angles, color distortions, or degrees of blurriness, diverse views are consistently derived from a common source object. This relationship implies statistical dependence, which is essential for the purpose of self-supervised learning. Nonetheless, the FMCA alone might not suffice, but a hierarchical adaptation is required.

Consider an source, unaugmented image $X \sim \mathbb{P}(\mathbf{X})$. Typically, self-supervised learning utilizes a stream of images produced through diverse augmentation protocols. This is often formulated as a transformation function $\mathcal{T}(X;v)$, accepting an image $X \in \mathcal{X}$ and a positive integer index $v \in \mathbb{Z}^+$, where each index $v$ symbolizes a specific augmentation protocol.

Aligning with the training procedure, we introduce a series of $L$ i.i.d. categorical random variables, $\mathbf{V} = \{\mathbf{v}_1, \cdots, \mathbf{v}_L\}$, denoting the execution of $L$ augmentations. In training, we sample an image $X \sim \mathbb{P}(\mathbf{X})$, and a set of indices $\{v_1, \cdots, v_L\} \sim \mathbb{P}(\mathbf{V})$. This samples $L$ views of the images, $\mathcal{T}(X;V) = \{\mathcal{T}(X;v), v\in {v_1, \cdots, v_L}\}$, perceived as a sequence of images representing their common source object under various angles, lighting conditions, or color variations.

Our proposition is that dependence arises between each individual view $\mathcal{T}(X;v)$, and its associated collection of multiple views $\mathcal{T}(X;V)$, where $\mathcal{T}(X;V)$ acts as a representation of the source object. This relationship across two scales, reflecting a hierarchy, requires a new formulation of dependence between a composite entity and its individual components. For this purpose, we propose a new joint distribution operating on two hierarchical levels.
\begin{definition}Consider a random process $\mathbf{X}^{(2)} = \{\mathbf{X}^{(2)}(l), l = 1,\cdots, L\}$, composed of $L$ smaller processes that all share the same support. We denote these smaller processes as $\mathbf{X}^{(1)}$. In essence, $\mathbf{X}^{(2)}$ and $\mathbf{X}^{(1)}$ symbolize two hierarchical levels, with $\mathbf{X}^{(2)}$ and $\mathbf{X}^{(1)}$ corresponding to the higher and lower levels, respectively. The lower-level marginal distribution $p(\mathbf{X}^{(1)})$ can be determined by collecting all possible realizations from the lower-level components, independent of $\mathbf{X}^{(2)}$, as $p(\mathbf{X}^{(1)} = {X}^{(1)}) = \frac{1}{L}\sum_{l=1}^L p(\mathbf{X}^{(2)}(l) = {X}^{(1)})$. Next, for a given higher-level realization ${X}^{(2)}$ of $\mathbf{X}^{(2)}$, the conditional distribution of its components is characterized by an empirical distribution:
\begin{equation}
p( {X}^{(1)} | \mathbf{X}^{(2)} = {X}^{(2)}) = \frac{1}{L} \sum_{l=1}^{L} \mathbbm{1} \{{X}^{(1)} \!=\! {X}^{(2)}(l)\}.
\label{distribution_def}
\end{equation}
This induces a cross-scale joint distribution $p_H({X}^{(1)}, {X}^{(2)})= p( {X}^{(1)} | {X}^{(2)}) p({X}^{(2)})$. Combine the joint and the marginal product, we define the induced density ratio $\rho_H({X}^{(1)}, {X}^{(2)}) = \frac{p({X}^{(1)}, {X}^{(2)})}{p({X}^{(1)})p({X}^{(2)})}$.
\label{definition1}
\end{definition}
\vspace{-5pt}
Definition~\ref{definition1} characterizes dependencies between hierarchical levels $\mathbf{X}^{(1)}$ and $\mathbf{X}^{(2)}$, where $\mathbf{X}^{(2)}$ is composed of $\mathbf{X}^{(1)}$. It outlines the statistical relationship between individual views $\mathcal{T}(X;v)$ and their associated collection $\mathcal{T}(X;V)$. Elaborating further on Definition~\ref{definition1}, sampling from marginal products permits independent samples from either level without considering their correspondence. Conversely, sampling from the joint requires sampling a $X^{(2)}$ from the higher level first, followed by sampling a $X^{(1)}$ from its components within $X^{(2)}$, which induces dependence. Our proposed density ratio contains this dependence information across two hierarchical levels.



With the density ratio established, we apply FMCA. Consistent with the theory (Eq.~\eqref{theory_decompose}), a spectral decomposition of the density ratio $\rho_H$ exists, and $\rho_H = \sum_{k=1}^\infty \sqrt{\sigma_k} {\phi_k} (X^{(1)}) {\psi_k} (X^{(2)})$, with orthonormal bases w.r.t. the marginal distributions. To approximate these bases, we employ two multivariate nonlinear mappers. The first mapper $\mathbf{L1}$ maps $X^{(1)}$ into a $K$-dimensional feature space, working with samples from the lower hierarchical level. The second mapper, denoted as $\mathbf{L2}$, working with $X^{(2)}$, maps the concatenation of $L$ feature maps, as generated by $\mathbf{L2}$, into another $K$-dimensional feature space corresponding to the higher level. These two nonlinear mappers $\mathbf{L1}$ and $\mathbf{L2}$, can essentially be used as the function approximators, denoted as $\mathbf{f}_\theta^{(1)}$ and $\mathbf{f}_\theta^{(2)}$, for approximating the basis functions of the density ratio decomposition. This inspires the Hierarchical Functional Maximal Correlation Algorithm (HFMCA).




\begin{prop}
 We denote the feature maps produced for the low-level and high-level hierarchies as $\mathbf{Z}^{(1)}$ and $\mathbf{Z}^{(2)}$, respectively, and represent the feature maps generated by the $L$ components of $\mathbf{Z}^{(2)}$ as ${\mathbf{Z}^{(1)}_l, l=1,\cdots, L}$. HFMCA solves the following optimization problem:
 \vspace{-3.5pt}
\begin{equation}
\resizebox{.85\linewidth}{!}{
$\begin{gathered}
\mathbf{R}_1 = \mathbb{E}[\mathbf{Z}^{(1)}  \mathbf{Z}^{(1)}{}^\intercal],\; \mathbf{R}_2 = \mathbb{E}[\mathbf{Z}^{(2)}  \mathbf{Z}^{(2)}{}^\intercal],\;
\mathbf{P}_{1, 2} = \frac{1}{L}\mathbb{E}[\sum_{l=1}^L\mathbf{Z}^{(1)}_l  \mathbf{Z}^{(2)}{}^\intercal ],\; \mathbf{R}_{1, 2} = \begin{bmatrix}\mathbf{R}_1 & \mathbf{P}_{1, 2} \\
\mathbf{P}^\intercal_{1, 2} & \mathbf{R}_2
\end{bmatrix},\\
\min_{\theta} r_{H} := \log\det \mathbf{R}_{1, 2} -\log \det \mathbf{R}_{1} - \log\det \mathbf{R}_{2}.
\end{gathered}$}
\label{hfmca_def}
\end{equation}
By the theory of FMCA, the optimal objective function reaches the leading eigenvalues of the density ratio $\rho_H$, with the neural networks reaching the leading orthonormal basis functions.
\label{proposition_2}
\end{prop}

This proposition lays the foundation for this paper, which involves constructing cost functions $r_H(\mathbf{Z}^{(1)}, \mathbf{Z}^{(2)})$ and employing the HFMCA for minimization. This procedure measures statistical dependence between two hierarchical levels and, most importantly, extracts multivariate features that are theoretically orthonormal.

\noindent \textbf{External costs for multiview systems.}  We propose to minimize the cost $r_H(\mathcal{T}(\mathbf{X};\mathbf{v}), \mathcal{T}(\mathbf{X};\mathbf{V}))$ using HFMCA to learn diverse features for self-supervised learning. Unlike conventional methods optimizing similarity measures between augmentation pairs, HFMCA employs an additional network after the backbone. The backbone CNN is first applied to the $L$ augmentations of a source image, extracting $L$ feature maps $\mathbf{Z}^{(1)}_l$ of $K$ dimensions. These $L$ feature maps are then concatenated in the feature channel, serving as inputs to the additional network and yielding the higher-level features $\mathbf{Z}^{(2)}$. Minimizing the log-determinants of marginal ACFs, $\mathbf{R}_1$ and $\mathbf{R}_2$, ensures orthonormality. Meanwhile, maximizing the joint ACF $\mathbf{R}_{1, 2}$ align the two sets of bases as parallel as possible. Solving this min-max problem effectively extracts shared information between two levels. HFMCA offers orthonormal features as the density ratio's basis functions, ensuring diversity. We refer to this as the external cost, reflecting dependencies from external knowledge or supervision.

\vspace{-1pt}

\section{Image Hierarchies: Modeling Internal Dependencies}
\vspace{-1pt}

\begin{figure}[t]
    \centering
    \includegraphics[width=.9\textwidth]{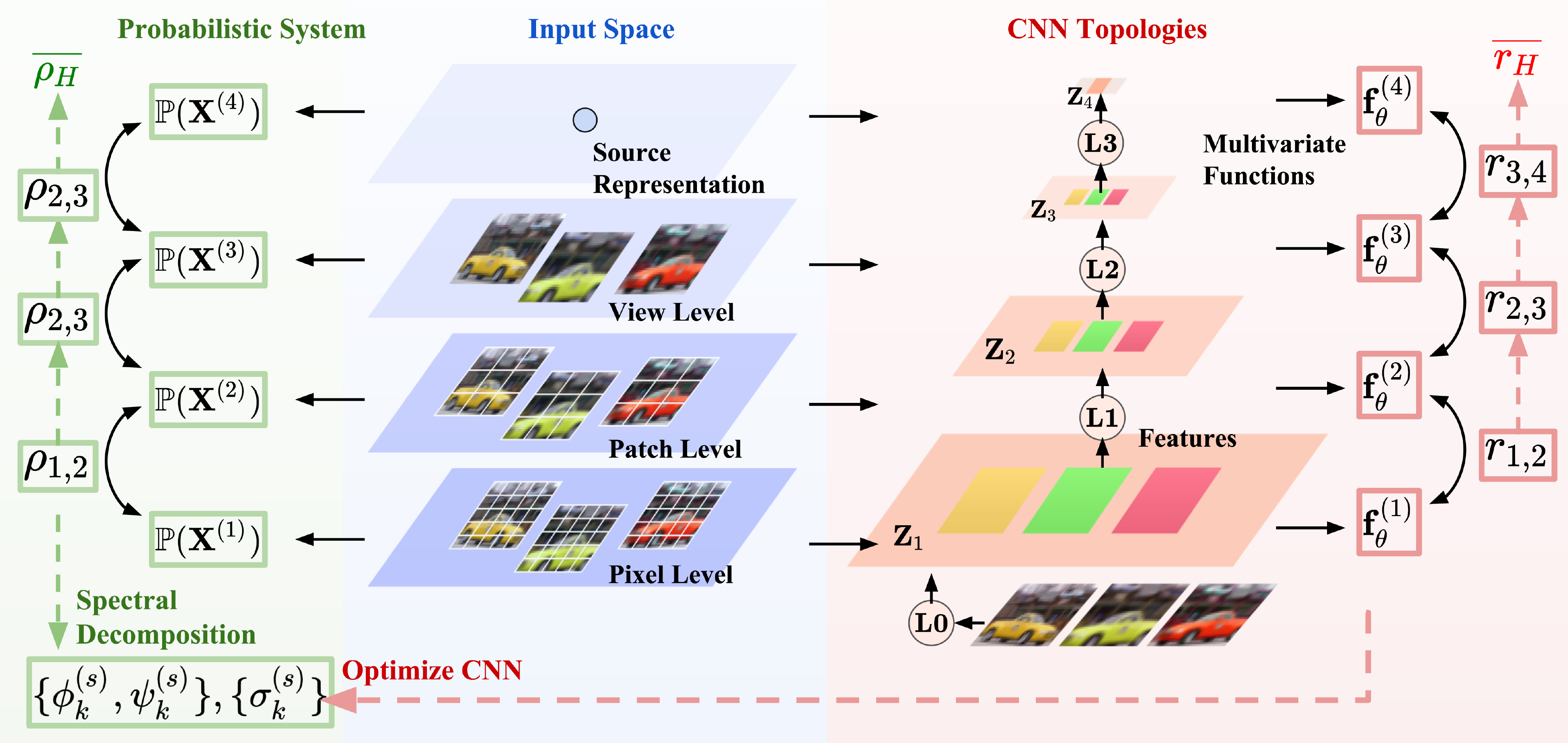}
    \caption{\small Overview of HFMCA considering both external and internal costs. It comprises three components: {\setlength{\fboxsep}{0pt}\colorbox{green1!15}{Probabilistic system}}, {\setlength{\fboxsep}{0pt}\colorbox{blue!15}{Input space}}, and {\setlength{\fboxsep}{0pt}\colorbox{red1!15}{CNN topology}}. {\setlength{\fboxsep}{0pt}\colorbox{blue!15}{Input space}} contains pixels, patches, and image augmentations (views), and their collection (source representation). {\setlength{\fboxsep}{0pt}\colorbox{green1!15}{Probabilistic system}} defines density ratios $\rho_{s, s+1}$ between neighboring scales, exhibiting the telescoping property to form the global density ratio $\overline{\rho_H}$. Each density ratio $\rho_{s, s+1}$ yields a spectral decomposition into bases $\{{\phi_k^{(s)}}, {\psi_k^{(s)}}\}$ and spectrum $\{\sigma_k^{(s)}\}$. {\setlength{\fboxsep}{0pt}\colorbox{red1!15}{CNN topology}} applies downsampling layers $\mathbf{L0}, \mathbf{L1}, \mathbf{L2}, \cdots$, each layer functioning as a universal mapper $\mathbf{f}_\theta^{(s)}$ of the receptive field, measurable against $\mathbb{P}(\mathbf{X}^{(s)})$. Cost functions $r_{s, s+1}$ optimize eigenvalues until identifying leading eigenvalues and bases. Spectrum (dependence measure) and bases (features) are acquired from training CNNs. External costs apply at the top scale, when multiple views compose  their respective collections. Internal costs apply at scales within image hierarchies, such as pixels, patches, and full images.} 
    \label{full_diagram_first}
\end{figure}

Empirical evidence from our experiments indicates that the use of HFMCA for self-supervision significantly accelerates training and enhances stability. To further interpret our approach, we must investigate the internal structures of images. Fig.~\ref{full_diagram_first} presents the full diagram of our proposed approach.

\subsection{Hierarchy in Focus: Pixels, Patches \& Images}

As multiple views represent one source object, a similar relationship is naturally present within image hierarchies between pixels, patches, and full images. Moreover, our construction also suggests that this relationship exists at local levels. For instance, the composition from pixels to patches is unaffected by and remains independent of how patches compose the full images. Thus, the most precise formulation is based on neighboring hierarchical levels.

We define a sequence of r.p. $\{\mathbf{X}^{(1)}, \mathbf{X}^{(2)}, \cdots, \mathbf{X}^{(S)}\}$, where $S$ denotes the number of hierarchical scales. In this context, $\mathbf{X}^{(1)}$ corresponds to pixels, $\mathbf{X}^{(S)}$ to images, and the intermediate processes $\mathbf{X}^{(2)}, \cdots, \mathbf{X}^{(S-1)}$ to image patches. Each element $\mathbf{X}^{(s)}$ in this sequence has spatial dimensions $(H_s, W_s)$ and is composed of pixels: $\mathbf{X}^{(s)} = \{{\mathbf{X}^{(s)}(i, j) \mid 1 \leq i \leq H_s, \ 1 \leq j \leq W_s }\}$. 

Our aim is to replicate the previous procedure to define the joint distribution across two neighboring hierarchical scales. This aligns with the standard patch creation procedure. Given a patch $X^{(s+1)}$ at scale $s+1$, it can be divided into multiple subpatches at a lower scale $s$. When two scales have nearly identical dimensions, the conditional distribution $p(\mathbf{X}^{(s)} = X^{(s)}|X^{(s+1)})$ has discrete, finite support of subpatches at scale $s$, modeled by an empirical distribution. Denote the differences in their patch dimensions as $\Delta H_s = H_{s+1} - H_s + 1$ and $\Delta W_s = W_{s+1} - W_s + 1$, we define
\vspace{-1pt}
\begin{equation}
\begin{aligned}
p( X^{(s)}|X^{(s+1)}) = \frac{1}{\Delta H_s \Delta W_s} \sum_{m=1}^{\Delta H_s} \sum_{n=1}^{\Delta W_s} \mathbbm{1} \{X^{(s)} \!=\! {X^{(s+1)}(m\!:m+H_s, n\!:n+W_s)}\}.
\label{joint_distribution}
\end{aligned}
\end{equation}
This induces a joint distribution $p_H(X^{(s)}, X^{(s+1)}) = p( X^{(s)}|X^{(s+1)}) p(X^{(s+1)})$ for every $s$. To sample from $p(X^{(s)}, X^{(s+1)})$, we first draw a $X^{(s+1)}$ at scale $s+1$, followed by sampling a subpatch $X^{(s)}$ within it, which maintains their dependence. The joint distribution $p_H(X^{(s)}, X^{(s+1)})$ further induces a series of density ratios $\rho_H(X^{(s)}, X^{(s+1)})$ for every scale $s$. 

\noindent \textbf{Telescoping property of density ratios.} An important property of our construction is the local existence of dependencies within hierarchies. Pixels first constitute patches, regardless of how these patches eventually combine to form images. Likewise, patches make up images, regardless of the correspondence between augmentations and the source object. This inherent trait uncovers the telescoping nature of dependencies, which can be characterized formally with density ratios:
\begin{prop} The global-level dependence for the image hierarchical sequence $\{\mathbf{X}^{(1)}, \mathbf{X}^{(2)}, \cdots, \mathbf{X}^{(S)}\}$ exists locally between neighboring hierarchical levels, as 
\begin{equation}
\begin{aligned}
\log \frac{p(X^{(1)}, X^{(2)}, \cdots, X^{(S)})}{p(X^{(1)}) p(X^{(2)}) \cdots p(X^{(S)}) } = \sum_{s=1}^{S-1} \log\rho_H({X^{(s)}, X^{(s+1)}}) := \log \overline{\rho_H}({X^{(1)}, \cdots, X^{(S)}}).
\end{aligned}
\label{telescoping_relation}
\end{equation}
We name this relationship the telescoping property
of density ratios. 
\end{prop}
This property matches our discussion and addresses the necessity and sufficiency of modeling statistical dependence between neighboring hierarchical levels. It also reveals a crucial characteristic: the global-level dependence can be defined and modeled at local levels. 




\vspace{-1pt}

\subsection{Functional Maximal Correlation via CNNs: Theoretical Solutions}

We have defined a sequence of density ratios $p_H(X^{(s)}, X^{(s+1)})$, ranging from scales $s=1$ to $S-1$. The remaining step is to apply HFMCA for decomposition. This follows the procedure detailed in Proposition~\ref{proposition_2}, which includes the implementation of the cost $r_H$ and the subsequent 
minimization. Remarkedly, the topology for this optimization aligns perfectly with the structure of a vanilla CNN.

Our construction is based on the assumption that each element in the CNN feature map can act as a universal approximator for its corresponding receptive field. This mapping relationship aligns with the functions needed as basis functions in the optimization. 

\begin{wrapfigure}{r}{0.34\textwidth}
  \centering
\includegraphics[width=1\textwidth]{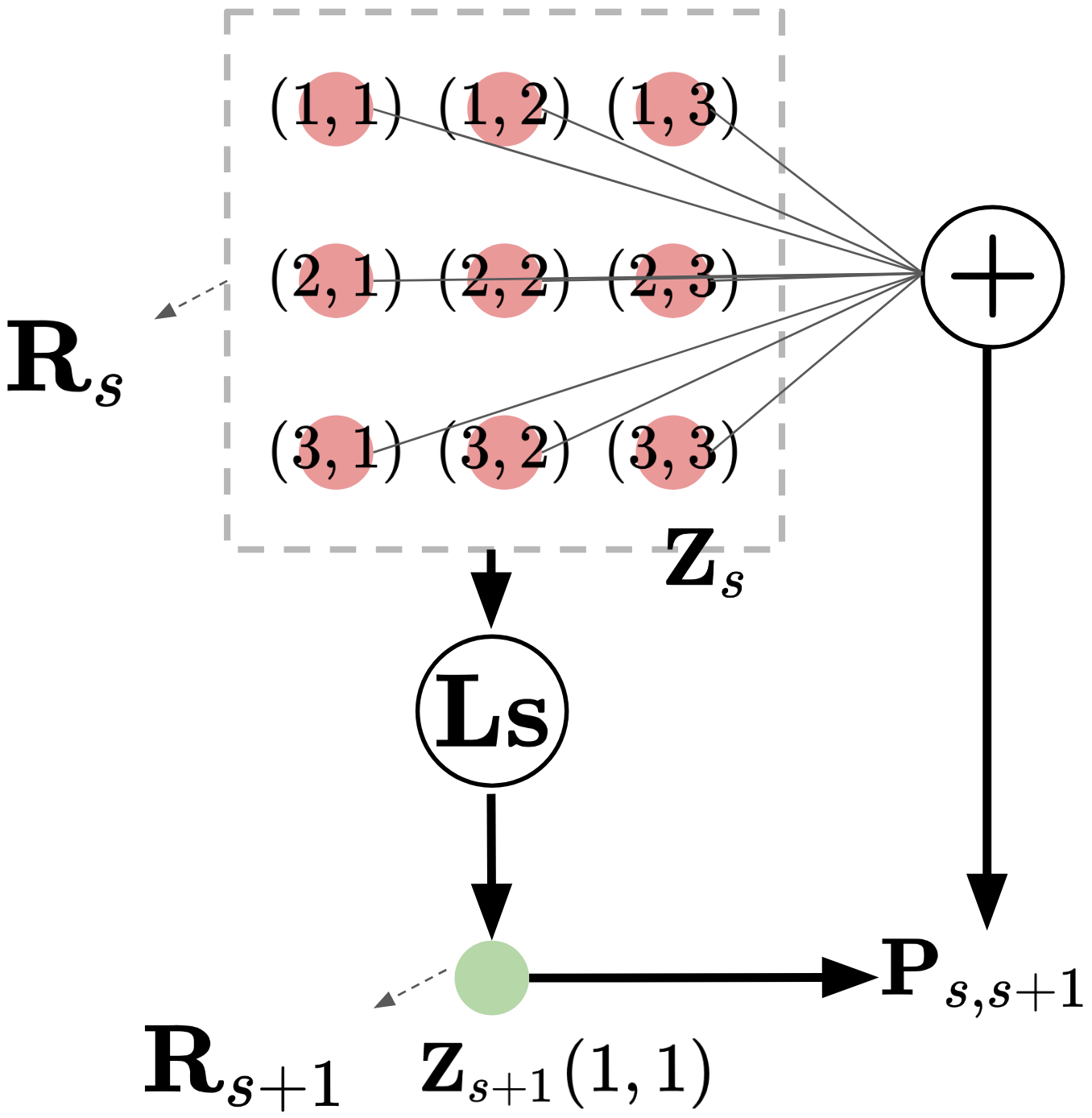}
  \caption{\small Illustration of HFMCA costs (Eq.~\eqref{hfmca_def}). Use region averages in $\mathbf{Z}_s$ and elements in $\mathbf{Z}_{s+1}$ to compute the CCF. Combine the CCF with ACFs of two marginals to construct cost functions $r_H$ at every layer $s$.
  \vspace{-18pt}
  }
  \label{figure2}
\end{wrapfigure}

As detailed in Proposition~\ref{proposition_2}, the execution of HFMCA involves first applying a feature network to each component patch, concatenating these lower-level features, and then passing them into another network to generate the higher-level features. This procedure bears a close resemblance to the convolution operation, where kernels are applied to subregions of the preceding layer's feature maps. We illustrate this with Fig.~\ref{figure2} and the following explanation.

Consider a CNN with $S$ layers. The initial layer consists of $1\times 1$ convolution kernels, yielding feature maps $\mathbf{Z}^{(1)}$, where the receptive fields correspond to individual pixels. Supposedly the second layer consists of $\Delta H_1 \times \Delta W_1$ convolution kernels (as refer to the patch size in Eq.~\eqref{joint_distribution}). This layer operates on the concatenation of $\Delta H_1 \times \Delta W_1$ elements from the feature maps of the first layer. We can infer that the receptive field of this second layer aligns with image patches of size $H_2 \times W_2$.

Hence, the receptive fields of first layer align with $\mathbf{X}^{(1)}$ in the hierarchical sequence, while those of the second layer align with $\mathbf{X}^{(2)}$. Following Proposition~\ref{proposition_2}, minimizing the cost $r_H$ between $\mathbf{Z}^{(1)}$ and $\mathbf{Z}^{(2)}$, denoted as $r_H(\mathbf{Z}^{(1)}, \mathbf{Z}^{(2)})$, effectively decomposes the density ratio between their receptive fields, $\rho_H(X^{(1)}, X^{(2)})$. 

Likewise, for the remaining layers with kernels dimensions $\Delta M_s \times \Delta N_s$, its receptive field will align precisely with the patch size of $\mathbf{X}^{(s)}$. By constructing and minimizing costs $r_H(\mathbf{Z}^{(s)}, \mathbf{Z}^{(s+1)})$ between neighboring CNN layers, we effectively decompose the density ratios $\rho_H(X^{(s)}, X^{(s+1)})$ between neighboring hierarchical levels.


\textbf{Internal costs for image hierarchies.} Suppose this CNN generates $S$ feature maps $\mathbf{Z}^{(1)},\cdots, \mathbf{Z}^{(S)}$. We introduce internal costs $r_H(\mathbf{Z}^{(s)}, \mathbf{Z}^{(s+1)})$ (per Proposition 2), and minimize the total cost $\min_\theta \sum_{s=1}^S r_H(\mathbf{Z}^{(s)}, \mathbf{Z}^{(s+1)})$. This minimization reveals statistical dependencies within image hierarchies. If an external cost $r_H(\mathbf{Z}^{(S)}, \mathbf{Z}^{(S+1)})$ is present, we formulate the task as follows,
\begin{equation}
\begin{aligned}
\min_\theta \; \lambda \sum_{s=1}^S r_H(\mathbf{Z}^{(s)}, \mathbf{Z}^{(s+1)}) +  r_H(\mathbf{Z}^{(S)}, \mathbf{Z}^{(S+1)}).
\end{aligned}
\end{equation}
Upon reaching the minimum, in accordance with the theory of FMCA (Eq.~\eqref{theory_decompose}), we derive an approximation of the sequence of density ratios $\rho_H(X^{(s)}, X^{(s+1)})$. This approximation yields a decomposition at each local level, expressed by the spectrum (a novel multivariate dependence measure), and the corresponding orthonormal bases at each layer. 
\section{Experiments}
Our experimental results focus on illustrating two primary strengths of HFMCA: 
 \begin{itemize}[leftmargin=*,topsep=-0.2em,itemsep=-0.4pt] 
 \item \textbf{External costs:} HFMCA exhibits faster convergence, higher accuracy, and improved stability in self-supervised learning with external costs. Our dependence measure can also serve as a quality indicator for the augmentation protocol, independently of classification accuracy.
 \item \textbf{Internal costs:}. Training with internal costs enhances interpretability. We observe a close resemblance between class supervision and self-supervision through a spectrum analysis. Moreover, visualization of the density ratio's telescoping property with HFMCA reveals internal representations within network layers, which help explain the obtained mappings during training.
 \end{itemize} 

 \noindent \textbf{Fast convergence in self-supervised learning.} Our HFMCA model, trained with external costs, exhibits faster convergence and superior accuracy in self-supervised learning, as shown in Table~\ref{table_performance}. We compared its performance with multiple benchmark models on CIFAR10 and CIFAR100, with the max accuracy achieved over $20$, $200$, and $800$ epochs reported, where HFMCA consistently outperformed them. All experiments use a consistent setup: a ResNet-18 backbone, batch size of $64$, SGD optimizer, a learning rate of $0.06$, and momentum of $0.9$, following benchmark settings. We use standard SimCLR protocols~\cite{chen2020simple} for augmentation and apply a KNN to embedded training images.



In HFMCA, for a batch of 64 images, we generate $128$-dimensional feature maps for $L=9$ distinct augmentations per image using a ResNet-18 backbone. These feature maps are then reshaped into a $3\times3$ grid, forming a tensor of size $(64, 128, 3, 3)$ which is fed into a $3$-layer CNN, creating a $128$-dimensional feature per source image. The external cost is constructed following Proposition~\ref{proposition_2} and Fig.~\ref{figure2}, while accuracy is evaluated via KNN on the final layer of the ResNet-18 backbone. We compare HFMCA with both internal and external costs (HFMCA-SS) and only with external costs (HFMCA-SS(*)). We find that HFMCA-SS(*) offers the fastest convergence and highest accuracy, while internal costs may slightly slow convergence without affecting accuracy.

Table~\ref{table_performance} highlights the benefits of shifting from the conventional similarity-contrastivity model to HFMCA's dependence-orthonormality framework. HFMCA promotes feature diversity via log-determinant-based cost functions, supported by orthonormal decompositions. Basically, it reformulates the task from contrasting two views to the measurement of statistical dependence among $L$ distinct views, and results in more efficient training.

\begin{table}[h]
    \centering
    \caption{\small Classification accuracy on CIFAR10 and CIFAR100 highlights HFMCA's effectiveness. HFMCA with external costs (HFMCA-SS(*)) converges fastest among methods, while HFMCA with both internal and external costs (HFMCA-SS) ranks second, retaining near-optimal accuracy.\vspace{-5pt}}
    \resizebox{.9\linewidth}{!}{%
\begin{tabular}{l c c c c c c c c}
    \toprule
    {\small \textbf{Method}} & {\small \textbf{Heads}} & \multicolumn{3}{c}{\small \textbf{CIFAR10}} & \multicolumn{3}{c}{\small \textbf{CIFAR100}}  \\
    \cmidrule(lr){3-5} \cmidrule(lr){6-8} 
        &  & Epoch 20 & Epoch 200 & Epoch 800 & Epoch 20 & Epoch 200 & Epoch 800 \\
        \midrule
        \textit{Methods with two views} \\
        MoCo~{\footnotesize \color{brown}~\cite{he2020momentum}}         & 128 & {57.2} & {83.8} & {90.0} & {22.3} & {45.7} & {69.8}  \\
        SimCLR~{\footnotesize \color{brown}~\cite{chen2020simple}}         & 128 & {46.7} & {82.2} & {87.5} & {19.6} & {43.9} & {65.7}  \\
        Barlow Twins~{\footnotesize \color{brown}~\cite{grill2020bootstrap}}  & 2048  & {45.7} & {83.5} & {85.7} & {28.1} & {47.1} & \textbf{70.9}  \\
        SimSiam~{\footnotesize \color{brown}~\cite{chen2021exploring}}        & 2048 & {50.5} & {83.7} & {90.0} & {22.5} & {39.9} & {66.0}  \\
        VICReg~{\footnotesize \color{brown}~\cite{wang2021vicreg}}          & 2048 & {44.8} & {81.2} & {90.2} & {20.3} & {37.8} & {68.5}  \\
        VICRegL~{\footnotesize \color{brown}~\cite{bardes2022vicregl}}        & 2048 & {43.2} & {78.7} & {89.7} & {21.5} & {41.2} & {67.3}  \\
        \midrule
        \textit{Methods with multiple views} \\
        FastSiam~{\footnotesize \color{brown}~\cite{pototzky2022fastsiam}} & 2048 & 76.8& 87.9& 90.1 & 45.8& {62.2}&69.9 \\
        \rowcolor{myblue!15}  HFMCA-SS(*)      & 128  & \textbf{81.8} & \textbf{89.3} & \textbf{90.7} & \textbf{47.5} & \textbf{67.6} & {70.3}  \\
        \rowcolor{myblue!15}  HFMCA-SS    & 128   & {79.3} & {85.7} & {90.1} & {43.3} & {65.1} & {67.9}  \\
        \bottomrule
    \end{tabular}}
    \label{table_performance}
\end{table} 



\noindent \textbf{Dependencies versus augmentation protocols.} HFMCA's strength as a statistic dependence measure is demonstrated through varying augmentation protocols. Our first observation is the impressive stability of HFMCA across all tests. We observe no occurrence of feature collapse, even under extreme augmentations described later. Second, the dependence  measurement provides a novel indication for evaluating the quality of augmentation protocols, independent of classification accuracy. 
 
Notably, even with strong augmentation, some level of dependence among views persists. We consistently observe a decrease in dependence level as we enrich the augmentation. The default augmentations for CIFAR10~\cite{chen2020simple} include random crops, color jitters, and gray scales. We test five distortion strengths across these three protocols. Each protocol is tested individually, keeping the other two at default values. Random crop strength varies from no cropping at all to the sampling and resizing of any patch from $1\times 1$ to $32\times 32$ as inputs. Color jitter strength refers to the intensity of distortions in brightness, contrast, saturation, and hue. Gray scale strength is the likelihood of images converting to gray scales, with the maximum strength making all images colorless.

\begin{figure}[b]
\includegraphics[width=0.9\linewidth]{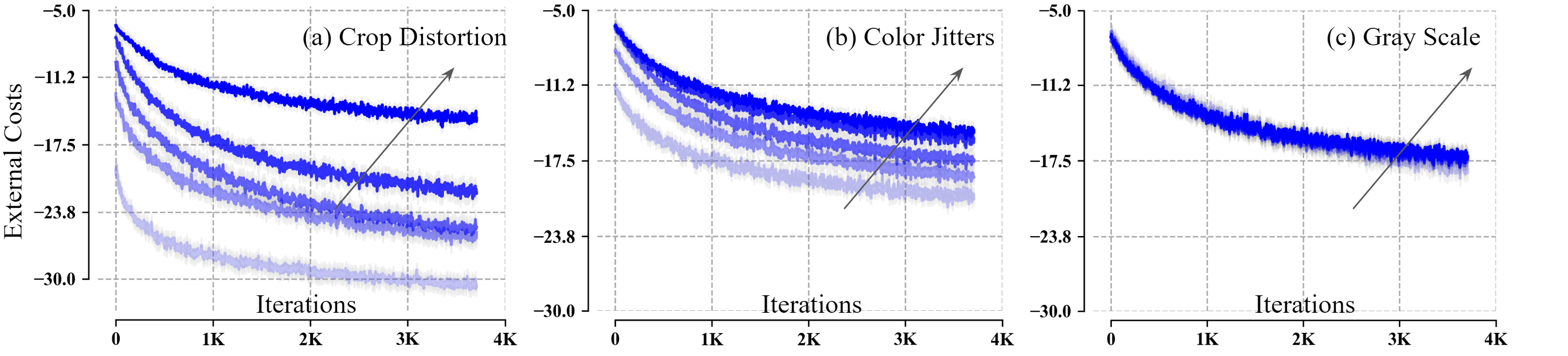}\vspace{-8pt}
\caption{\small The learning dynamics of external costs (dependence level) are displayed for five distortion strengths across three protocols. The arrow's direction indicates an increase in distortion strength. A lower cost value corresponds to higher dependence. The figure implies that as distortion strength increases, the dependence level decreases. However, even in extreme cases, a consistent level of dependence remains, intrinsic to the dataset. \vspace{-10pt}}
\label{learning_dynamics}
\end{figure}

Fig.~\ref{learning_dynamics} shows training dynamics of the external cost as the dependence level, indicating that random crops impact the most, followed by color jitters, and gray scale. Increased distortion strength reduces dependence level (increases costs), but never reaches strict independence. Intriguingly, even in extreme cases, the learning settles at a certain level of dependence intrinsic to the dataset, which can be interpreted as the intrinsic dimension of the data set. Modeling this intrinsic level of dependence, which is unaffected by the augmentation's richness, can be fundamental to self-supervised learning.

Table~\ref{table_aug} further supports our argument by showing the classification accuracy (A) and external costs (EC) for these experiments. The results further support HFMCA's robustness, showing no major accuracy drop or feature collapse with increased distortion. A decrease in external costs corresponds to an increase in classification accuracy. This consistency suggests our dependence measure's potential for evaluating the quality of different augmentation protocols.





\sidecaptionvpos{figure}{c}

\begin{table}[t]
\resizebox{.8\linewidth}{!}{
\centering
\caption{\small A comparison of classification accuracy (A) and external costs (EC) across three protocols. An increase in distortion strength leads to decreased dependence but enhances classification accuracy. The external costs never retain zero in all scenarios, suggesting an intrinsic level of dependence within the dataset. \vspace{-10pt}} 
\label{table_aug}
\begin{tabular}{c c c}
    \subcaptionbox{Crop Distortion}{
        \begin{tabular}{S[table-format=1.2] S[table-format=2.2] S[table-format=-2.2]}
        \toprule
         {Strength}& {A (\%)} & {EC} \\
        \midrule
        \rowcolor{myblue!0} 0 &  24.8 & -30.4 \\
        \rowcolor{myblue!5} 0.25 &  54.5& -26.1 \\
        \rowcolor{myblue!10} 0.5 & 67.7 & -25.4 \\
        \rowcolor{myblue!15} 0.75 & 71.2 & -21.9 \\
        \rowcolor{myblue!20} 1 & 69.9 &  -15.1\\
        \bottomrule
        \end{tabular}
        } &
    \subcaptionbox{Color Jitters}{
        \begin{tabular}{S[table-format=1.1] S[table-format=2.2] S[table-format=-2.2]}
        \toprule
        {Strength} & {A (\%)} & {EC} \\
        \midrule
        \rowcolor{myblue!0}  0 & 49.0 & -20.4 \\
        \rowcolor{myblue!5}0.25 & 69.6 & -18.8 \\
        \rowcolor{myblue!10} 0.5 & 71.0 & -17.5 \\
        \rowcolor{myblue!15}0.75 & 70.6 & -15.7 \\
        \rowcolor{myblue!20} 1 & 70.8 & -15.0 \\
        \bottomrule
        \end{tabular}
        } &
    \subcaptionbox{Grey Scale}{
        \begin{tabular}{S[table-format=1.2] S[table-format=2.2] S[table-format=-2.2]}
        \toprule
        {Strength} & {A (\%)} & {EC} \\
        \midrule
        \rowcolor{myblue!0} 0 & 61.2 & -18.1 \\
        \rowcolor{myblue!5}0.25 & 71.2 & -17.4 \\
        \rowcolor{myblue!10} 0.5 & 70.4 & -17.3 \\
        \rowcolor{myblue!15}0.75 & 71.4 & -17.2 \\
        \rowcolor{myblue!20} 1 & 70.7 & -17.1 \\
        \bottomrule
        \end{tabular}
        }\vspace{-10pt}
\end{tabular}}
\end{table}

\begin{figure}[b]
\vspace{-10pt}\includegraphics[width=0.85\linewidth]{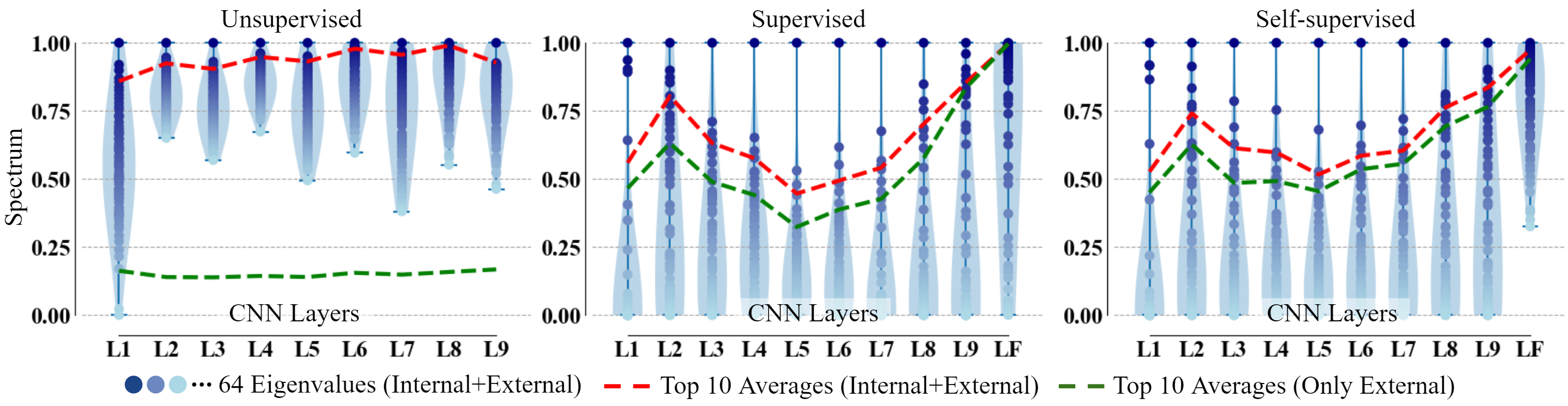}\vspace{-10pt}
\caption{\small Spectrum across layers under three supervision types. Each plot column represents the spectrum of a specific CNN layer, ranging from $\textbf{L1}$ to the last layer $\textbf{LF}$. Each spectrum consists of $64$ eigenvalues, described as blue dots. The dots' color intensity reflects their rank in the spectrum, with the blue background showing their distributions. In unsupervised scenarios, we consistently observe the presence of large eigenvalues, indicating the dataset's intrinsic dimensionality. Supervision and self-supervision significantly impact the eigenvalues in the middle layers, creating large null spaces to meet the goal of discrimination. \vspace{-10pt}}
\label{eigennalysis_cross_layer}
\end{figure}

\begin{figure}[t]
\includegraphics[width=0.85\linewidth]{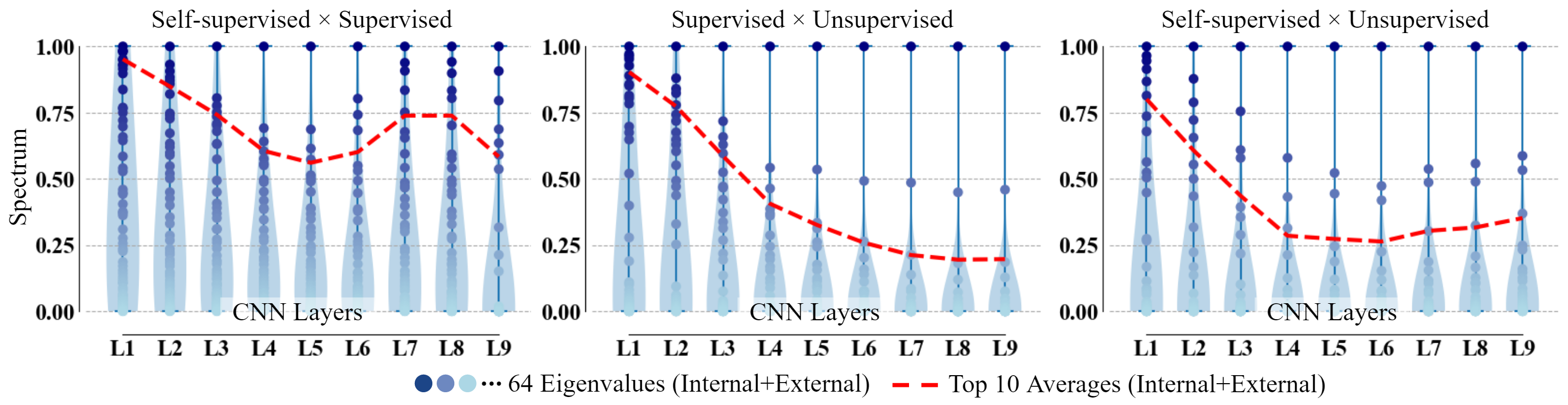}\vspace{-8pt}
\caption{\small Spectrum across pairs of supervision types. Eigenvalues, by projecting eigenfunctions onto each other, quantify the alignment between supervision pairs across network layers. Supervised and unsupervised learning exhibit different internal representations due to fewer large eigenvalues, particularly at the top layers. Supervised and self-supervised learning settings learn very similar spaces, albeit with some differences in the middle layers where most null space projections occur. \vspace{-8pt}} 
\label{eigennalysis_cross_method}
\end{figure}

\textbf{Explainability of internal representations for supervision \& self-supervision} Shifting our focus to representations, we investigate HFMCA's behavior under different supervision types using the learned spectrum for interpretability. Additional experiments illustrated in Fig.~\ref{eigennalysis_cross_layer} and Fig.~\ref{eigennalysis_cross_method}, conducted on CIFAR10, include unsupervised (where only internal costs are used) and supervised scenarios (where $L$ distinct views are substituted with $L$ samples from the same class), with and without internal costs. We use a modified CNN backbone and hyperparameters detailed in the appendix. The learned HFMCA eigenspectrum between pairs of layers is very telling. The eigenvalues range between $0$ and $1$ because of cost normalization and reflect dependence strength. Since dependence is associated with the correlation between the eigenfunctions, we explore this interpretation to discuss the layer effective dimension (different from the number of eigenfunctions that are kept at $64$). For the unsupervised case, excluding the first layer, the eigenvalues are basically in the same range, mostly above $0.5$, which means that there are minor modifications in the space dimensions due to the nonlinear mappings, but the eigenvalue distribution is always far from $0$. Hence, the dynamic range of the eigenvalues across layers shows that the dimensionality of the projection spaces  oscillates around the intrinsic dimensionality of the input data set. 

This picture changes drastically for the supervised and self-supervised cases, due to the external desired responses that force discrimination in the network input-output map. The dimensionality of the labels in the supervised case is much less than the intrinsic dimension of the input, and so we see a large number of eigenvalues close to zero (light blue regions), which means that the data is being projected to a smaller subspace across the layers. More importantly, notice that the higher density of zero eigenvalues occurs in the middle layers. For the self-supervised case, we have a similar picture, with a notable difference that the spread of eigenvalues in the last layer closely resembles that of the unsupervised case. This alignment is expected since the desired output is based on the source image itself. So we can conclude that the discrimination is affecting mostly the eigenvalues in the middle layers, creating large null spaces to meet the goal of classification. This explains why the feature collapse is so common in these types of applications. We also plot in a red dotted line the average of the $10$ largest eigenvalues that corroborate the analysis of the number of zero eigenvalues. 

The eigenvalues have another important application because they quantify the solid angle between eigenfunctions. This is particularly important when comparing the learned eigenfunctions across different supervision settings. Fig.~\ref{eigennalysis_cross_method} shows the eigenvalues between each pair of supervisions at each layer, by projecting eigenfunctions onto each other, to quantify the alignment of the spanned projection spaces (values close to $1$/$0$ mean parallel/orthogonal eigenfunctions, respectively). Notice that there are very few large eigenvalues across the supervised settings versus unsupervised, in particular on the top layers, meaning that their internal representations are quite different. The eigenvalues of the supervised versus self-supervised are much larger, meaning that they learn very similar spaces, particularly at the initial and final layers. This similarity decreases in the intermediate layers, which is precisely where most null space projections occur. This analysis provides a very specific understanding of the internal representations of complex networks across different settings, owing to the proposed methodology.

\begin{SCfigure}[][h]
  \centering
\includegraphics[width=.6\textwidth]{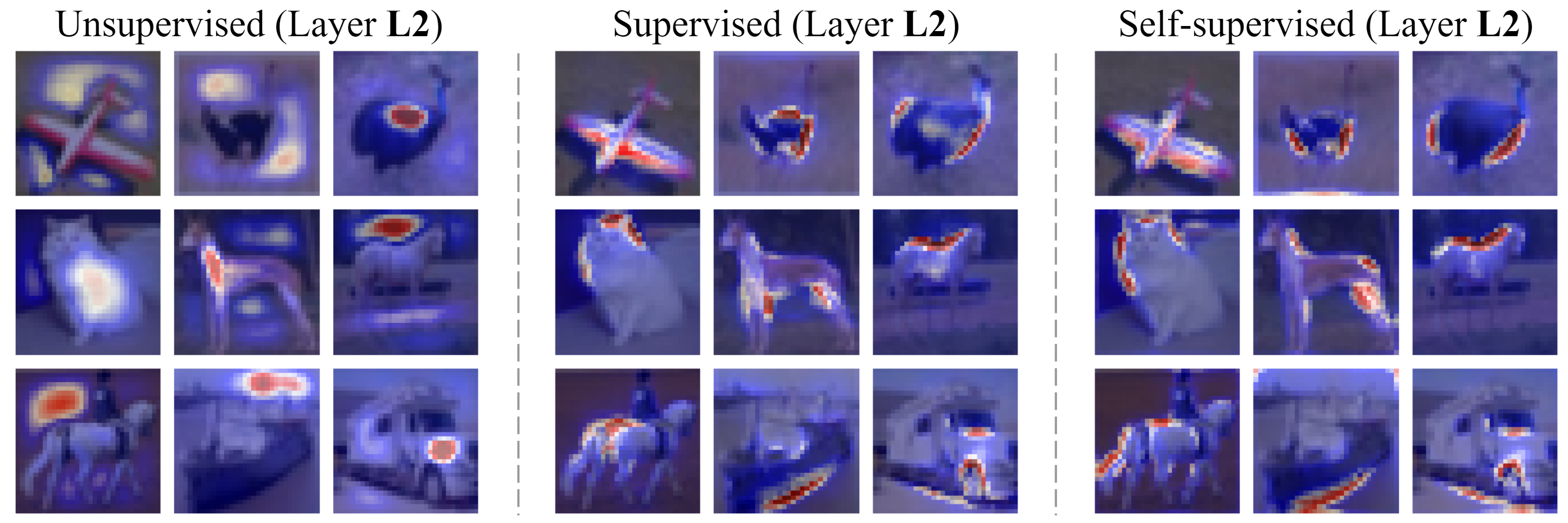}
  \caption{\small Global-to-local density ratio response of $9$ CIFAR10 samples for layer $\mathbf{L2}$. The boundaries of objects, and the interactions between different parts of an object, play a critical role in learning. This further demonstrates the resemblances between supervision and self-supervision.}
  \label{visualize_heat_map_multiple_samples}
\end{SCfigure}

\textbf{Visualizing telescoping density ratios.} An important component of our hierarchical dependence model is the telescoping property.  We demonstrate that the density ratios between two neighboring layers identify their dependencies, which effectively captures the global dependence information by extending to the entire network. Thus, starting from the top layer, we calculate the local density ratios between neighboring layers, passing these density ratios down to the bottom layers, layer by layer. Fig.~\ref{visualize_heat_map_multiple_samples} displays the local response at layer $\mathbf{L2}$ across three learning setups, revealing the most informative regions. We observe that the boundaries of objects, and the interactions between different parts of an object (such as how wheels are connected to the car body), play a critical role in learning. 

This process is quite similar to the effect of backpropagation of errors through gradients, but it is much more efficient and principled because it transmits statistical dependence instead of just the simple gradient of the error, which quantifies only the maximum rate of change. In our opinion, the remarkable  performance of our method in Table~\ref{table_performance} can be attributed to its telescoping property. Effectively, the telescoping property could serve as an alternative to error backpropagation training, potentially facilitating the effective layer-by-layer training of deep networks, and mitigating the issue of vanishing gradients while preserving interpretability on the image plane.

\vspace{-7pt}

\section{Discussion}

This paper proposes HFMCA as a method to achieve fast convergence and enhanced stability in self-supervised learning. HFMCA characterizes dependencies across hierarchical levels, further applied to image hierarchies via CNN training, enhancing feature interpretability. This paper explores the constituents of an image dataset's dependencies, including internal ones arising from pixel or patch alignment in the spatial domain and external ones derived from knowledge or supervision. Our study has not yet incorporated local-level supervision, such as patch augmentation~\cite{tong2023emp}, which can be explored in future work.

\end{document}


\maketitle

\vspace{-50pt}
\section{Pseudocode and Algorithm Details}

In this section, we will further illustrate the proposed methods, including the decomposition scheme with CNNs, obtaining basis functions and the spectrum through normalizations, and producing the local density ratio response.





\subsection{Details of Decomposition with CNNs}


Continuing from Section~4.2 of the paper, we further illustrate the use of CNN topologies in applying density ratio decomposition to image hierarchies.

\textbf{Notations.} To make the illustration more clear, we first clarify the notations that we use:
\vspace{-5pt}
 \begin{itemize}[leftmargin=*]
 \item $\mathcal{F}_\theta^{(1)}, \mathcal{F}_\theta^{(2)}, \cdots, \mathcal{F}_\theta^{(S)}$: Convolution layers of the CNN, each a function receiving patches from preceding feature maps.  \vspace{-2pt}
  \item ${\widehat{\mathcal{F}_\theta^{(1)}}}, {\widehat{\mathcal{F}_\theta^{(2)}}}, \cdots, {\widehat{\mathcal{F}_\theta^{(S)}}}$: Each CNN layer treated as a function of the corresponding receptive field in input images. \vspace{-1pt}
 \item $\mathbf{Z}_1, \mathbf{Z}_2, \cdots, \mathbf{Z}_S$:  Feature maps at each layer. \vspace{-1pt}
 \item $\mathbf{R}_\phi^{(s)}, \mathbf{R}_\psi^{(s)}, \mathbf{P}_{\phi, \psi}^{(s)}, \mathbf{R}_{\phi, \psi}^{(s)}, 1 \leq s \leq S$: Autocorrelation functions (ACFs) and crosscorrelation functions (CCFs) defining the optimization cost at each layer.
 \end{itemize} \vspace{-3pt}

\textbf{CNN as function approximators of feature maps.} Consider a CNN with $S$ layers. The first layer consists of $1\times 1$ convolution kernels (denote as $\mathcal{F}_\theta^{(1)}$). The remaining layers have kernels of dimensions $\Delta M_s \times \Delta N_s$ (denote as $\mathcal{F}_\theta^{(2)}, \cdots, \mathcal{F}_\theta^{(S)}$). We assume a fixed channel number $K$, and denote each layer's feature map, a tensor with dimensions $(H_s, W_s, K)$, as $\mathbf{Z}_s = \{{{\mathbf{Z}^{(s)}(i, j) \mid 1 \leq i \leq H_s, \ 1 \leq j \leq W_s }}\in \mathbb{R}^{H_s \times W_s \times K}\}$.

Every convolutional layer can be treated as a mapping function $\mathcal{F}_\theta^{(s)}: \Delta M_s \times \Delta N_s \rightarrow \mathbb{R}^K$, which operates on the outputs of its preceding layer, as $\mathbf{Z}_{s+1}(i, j) = \mathcal{F}_\theta^{(s+1)}(\mathbf{Z}_s(i:i+\Delta M_s, j:j+\Delta N_s))$. It follows that each element $\mathbf{Z}_{s}(i, j)$ in the feature map corresponds to a receptive field of dimensions $(H_s, W_s)$ in the input images, which we denote as $\mathbf{Z}_{s}(i, j) = \widehat{\mathcal{F}_\theta^{(s+1)}}(\mathbf{X}(i:i+H_s, j:j+W_s))$. This sequence of functions $\{\widehat{\mathcal{F}_\theta^{(s)}}:\mathcal{X}^{(s)} \rightarrow \mathbb{R}^K\}$ can be directly used as basis functions for density ratio decomposition proposed in Proposition 2 of the main paper. 

\textbf{Minimization criterion.} Taking into account the defined forms of joint distributions, the ACFs and CCFs can be defined at each layer and between two neighboring layers, taking the following forms:
\begin{equation}
\resizebox{.88\linewidth}{!}{
$\begin{gathered}
\mathbf{R}_\phi^{(s)}  = \frac{1}{M_\phi^{(s)}}\mathbb{E}[\sum_{i, j} \sum_{i'=i}^{i+\Delta M_s} \sum_{j'=j}^{j+\Delta N_s} \mathbf{Z}_s(i', j')\mathbf{Z}_s^\intercal(i', j')], \;\mathbf{R}_\psi^{(s)} = \frac{1}{M_\psi^{(s)}}\mathbb{E}[\sum_{i, j}\mathbf{Z}_{s+1}(i, j)\mathbf{Z}_{s+1}^\intercal(i, j)]\\
\mathbf{P}_{\phi, \psi}^{(s)} = \frac{1}{M_\phi^{(s)}} \mathbb{E}[\sum_{i, j} \sum_{i'=i}^{i+\Delta M_s} \sum_{j'=j}^{j+\Delta N_s} \mathbf{Z}_s(i', j') \mathbf{Z}_{s+1}^\intercal(i, j)], \mathbf{R}_{\phi, \psi}^{(s)} = \begin{bmatrix} \mathbf{R}_\phi^{(s)} & \mathbf{P}_{\phi, \psi}^{(s)} \\ \mathbf{P}_{\phi, \psi}^{(s)\intercal} & \mathbf{R}_\psi^{(s)}
\end{bmatrix},
\end{gathered}$}
\label{implementaiton_RP}
\end{equation}
where ${M_\phi^{(s)}}$ and ${M_\psi^{(s)}}$ simply denote the count of additions. The definition of $\mathbf{R}_\phi^{(s)}$ arises from how each layer is applied to the boundaries. For example, only one element in $\mathbf{Z}_{s+1}$ has a mapping relationship with the boundaries of $\mathbf{Z}_{s}$. Therefore, we examine each element in $\mathbf{Z}_{s+1}$, identify its corresponding inputs in $\mathbf{Z}_{s}$, compute the ACFs for all such related elements, and repeat this process for every element in $\mathbf{Z}_{s+1}$.

The internal costs and the minimization task can be written as
\begin{equation}
\resizebox{.95\linewidth}{!}{
$\begin{gathered}
r_{s, s+1} = \log\det \mathbf{R}_{\phi, \psi}^{(s)} -  \log\det \mathbf{R}_{\phi}^{(s)} - \log\det \mathbf{R}_{\psi}^{(s)}, \; 1 \leq s\leq S-1;\;\; \overline{r} = \sum_{s=1}^{S-1} r_{s, s+1}; \;\; \operatorname*{minimize}_{\theta} \;\overline{r}.
\end{gathered}$}
\label{minimization_problem}
\end{equation}
We provide the pseudocode for optimizing the internal costs in Algorithm~\ref{algorithm2}. 

\begin{algorithm}
\caption{HFMCA - Internal Costs}
\begin{algorithmic}[1] 
\STATE Initialize CNN with downsampling blocks $\mathcal{F}_\theta^{(1)}, \mathcal{F}_\theta^{(2)}, \mathcal{F}_\theta^{(3)}, \cdots, \mathcal{F}_\theta^{(S)}$
\WHILE{convergence is not reached}
\STATE Sample a batch of images $\mathbf{X}$; $\mathbf{Z}_1 = \mathcal{F}_\theta^{(1)}(\mathbf{X})$; $\overline{r}$ = 0
\FOR{$s = 1, \cdots, S-1$}
\STATE $\mathbf{Z}_{s+1} = \mathcal{F}_\theta^{(s+1)}(\mathbf{Z}_{s})$
\STATE $\mathbf{R}_\psi^{(s)} = \text{mean}(\mathbf{Z}_{s+1}(i, j)\mathbf{Z}_{s+1}^\intercal(i, j))$
\STATE For every $\mathbf{Z}_{s+1}^\intercal(i, j)$, find its receptive field in $\mathbf{Z}_{s}$: $\{\mathbf{Z}_{s}(i',j'), i'\in I, j' \in J\}$
\STATE $\mathbf{R}_\phi^{(s)}  = \text{mean}\left((\frac{1}{|I||J|}\sum_{i'}\sum_{j'}\mathbf{Z}_{s}(i',j')\mathbf{Z}_{s}^\intercal(i',j')\right)$
\STATE $\mathbf{P}_{\phi, \psi}^{(s)} = \text{mean}\left(\frac{1}{|I||J|}\sum_{i'}\sum_{j'}\mathbf{Z}_{s}(i',j') \mathbf{Z}_{s+1}^\intercal(i,j) \right)$
\STATE $\mathbf{R}_{\phi, \psi}^{(s)} = \mathbf{R}_{\phi, \psi}^{(S)} = \begin{bmatrix} \mathbf{R}_\phi^{(S)} & \mathbf{P}_{\phi, \psi}^{(S)} \\ \mathbf{P}_{\phi, \psi}^{(S)\intercal} & \mathbf{R}_\psi^{(S)}
\end{bmatrix}$
\STATE $r_{s, s+1} = \log\det \mathbf{R}_{\phi, \psi}^{(s)} - \log\det \mathbf{R}_\phi^{(s)} - \log\det \mathbf{R}_\psi^{(s)}$; $\overline{r} \leftarrow \overline{r} + r_{s, s+1}$
\ENDFOR
\STATE SGD: $\theta \leftarrow\theta + {\partial \overline{r}}/{\partial \theta}$
\ENDWHILE



\end{algorithmic}
\label{algorithm2}
\end{algorithm}

\textbf{Gradient estimation.} In our implementation, an adaptive filter can be added for gradient estimation, similar to a conventional Adam optimizer~\cite{kingma2014adam}. Note that the gradient of $r_{s, s+1}$ has the form
\begin{equation}
\begin{gathered}
\frac{\partial r_{s, s+1}}{\partial \theta} = \textbf{Tr} ((\mathbf{R}_{\phi, \psi}^{(s)})^{-1}\frac{\partial \mathbf{R}_{\phi, \psi}^{(s)}}{\partial \theta}) - \textbf{Tr} ((\mathbf{R}_{\phi}^{(s)})^{-1}\frac{\partial \mathbf{R}_{\phi}^{(s)}}{\partial \theta}) - \textbf{Tr} ((\mathbf{R}_{\psi}^{(s)})^{-1}\frac{\partial \mathbf{R}_{\psi}^{(s)}}{\partial \theta}).
\end{gathered}
\end{equation}
Thus, we use adaptive filters to estimate the three ACFs, and substitute the argument within the inverse function with these estimated values. We provide the pseudocode for this procedure in Algorithm~\ref{algorithm3}.

\begin{algorithm}
\caption{Adaptive Filters for Gradient Estimation}
\begin{algorithmic}[1] 
\STATE $k=1$; Initiate ACFs/CCFs estimators $\{\Tilde{\mathbf{R}}_\phi^{(s)}$, $\Tilde{\mathbf{R}}_\psi^{(s)}, \Tilde{\mathbf{R}}_{\phi, \psi}^{(s)}\}_{s=1}^S$
\WHILE{convergence is not reached}
\FOR{$s = 1, \cdots, S$}
\STATE $\Tilde{\mathbf{R}}_\phi^{(s)} \leftarrow \beta \Tilde{\mathbf{R}}_\phi^{(s)} +  (1-\beta) {\mathbf{R}}_\phi^{(s)}$ 
\STATE $\Tilde{\mathbf{R}}_\psi^{(s)} \leftarrow \beta \Tilde{\mathbf{R}}_\psi^{(s)} +  (1-\beta) {\mathbf{R}}_\psi^{(s)}$ 
\STATE $\Tilde{\mathbf{R}}_{\phi, \psi}^{(s)} \leftarrow \beta \Tilde{\mathbf{R}}_{\phi, \psi}^{(s)} +  (1-\beta) {\mathbf{R}}_{\phi, \psi}^{(s)}$ 
\STATE $\widehat{\mathbf{R}}_\phi^{(s)} = \Tilde{\mathbf{R}}_\phi^{(s)}/(1-\beta^k)$; $\widehat{\mathbf{R}}_\psi^{(s)} = \Tilde{\mathbf{R}}_\psi^{(s)}/(1-\beta^k)$; $\widehat{\mathbf{R}}_{\phi, \psi}^{(s)} = \Tilde{\mathbf{R}}_{\phi, \psi}^{(s)}/(1-\beta^k)$
\STATE Estimate gradients: 
\STATE  \hspace{0.2cm} {\large $\frac{\partial r_{s, s+1}}{\partial \theta}$} $\approx \text{pinv}(\widehat{\mathbf{R}}_{\phi, \psi}^{(s)})$ {\large $\frac{\partial \mathbf{R}_{\phi, \psi}^{(s)}}{\partial \theta}$} - $\text{pinv}(\widehat{\mathbf{R}}_{\phi}^{(s)})$ {\large $\frac{\partial \mathbf{R}_{\phi}^{(s)}}{\partial \theta}$} - $\text{pinv}(\widehat{\mathbf{R}}_{ \psi}^{(s)})$ {\large $\frac{\partial \mathbf{R}_{ \psi}^{(s)}}{\partial \theta}$}
\ENDFOR
\STATE $k\leftarrow k+1$
\ENDWHILE 
\end{algorithmic}
\label{algorithm3}
\end{algorithm}

\subsection{Pseudocode for HFMCA Self-supervised Learning}
Continuing from Section~3 of the paper, we provide the pseudocode for HFMCA with self-supervision in Algorithm~\ref{algorithm1}. 

\begin{algorithm}
\caption{HFMCA - External Costs}
\begin{algorithmic}[1] 
\STATE Choose number of views $L$; Initialize CNN $\mathcal{F}_\theta^{(1)}, \cdots, \mathcal{F}_\theta^{(S)}$; Initialize $\mathcal{F}_\theta^{(S+1)}:K\times L \rightarrow \mathbb{R}$
\WHILE{convergence is not reached}
\STATE Sample a batch of images $\mathbf{X}$
\STATE \textbf{if self-supervised}:
    \STATE \hspace{0.2cm} Create $L$ augmentations of $\mathbf{X}$, resulting in $\mathbf{X}_1, \mathbf{X}_2, \dots, \mathbf{X}_L$
\STATE \textbf{else if supervised}:
    \STATE \hspace{0.2cm} Sample $L-1$ batches with samples of the same class as $\mathbf{X}$, resulting in $\mathbf{X}_1, \mathbf{X}_2, \dots, \mathbf{X}_L$
\STATE  $\mathbf{Z}_{1, l} = \mathcal{F}_\theta^{(1)}(\mathbf{X}_l)$ for all $l$ ; $\overline{r}$ = 0
\FOR{$s = 1, \cdots, S-1$}
\STATE $\mathbf{Z}_{s+1, l} = \mathcal{F}_\theta^{(s+1)}(\mathbf{Z}_{s, l})$ for all $l$
\STATE \textbf{if internal costs}: Build $r_{s, s+1}$; $\overline{r} \leftarrow \overline{r} + r_{s, s+1}$
\ENDFOR
\STATE Concatenate feature maps as inputs: $\mathbf{Z}_{S+1} = \mathcal{F}_\theta^{(S+1)}([\mathbf{Z}_{S, 1}, \mathbf{Z}_{S, 2}, \cdots, \mathbf{Z}_{S, L}]^\intercal)$
\STATE $r_{S, S+1} = \text{mean} \left(\frac{1}{L}\sum_{l=1}^L \mathbf{Z}_{S, l} \mathbf{Z}_{S, L+1}^\intercal \right)$; \;$\overline{r} \leftarrow \overline{r} + r_{S, S+1}$ 
\STATE SGD: $\theta \leftarrow\theta + {\partial \overline{r}}/{\partial \theta}$
\ENDWHILE
\end{algorithmic}
\label{algorithm1}
\end{algorithm}

\subsection{Retrieve Eigenspectrum and Basis Functions.}

After training, we apply a standard normalization scheme to obtain the spectrum and basis functions from neural network outputs. This includes the first step by enforcing orthonormality with $\overline{\phi_\theta^{(s)}} = {\mathbf{R}_\phi^{(s)-\frac{1}{2}}} \widehat{\mathcal{F}_\theta^{(s)}}, \overline{\psi_\theta^{(s)}} = {\mathbf{R}_\psi^{(s)-\frac{1}{2}}}\widehat{\mathcal{F}_\theta^{(s+1)}}$. The second step is to apply the singular-value decomposition such that the functions are invariant to the conditional mean operator, which follows
\begin{equation}
\begin{gathered}
\mathbb{E}[\overline{\phi_\theta^{(s)}}\;\overline{\psi_\theta^{(s)}}{}^\intercal]  = \mathbf{U}_s \widehat{\mathbf{\Sigma}_s} {}^{\frac{1}{2}} \mathbf{V}_s^\intercal,\; \widehat{\mathbf{\Sigma}_s} = \text{diag}([\widehat{\sigma_1^{(s)}}, \cdots, \widehat{\sigma_K^{(s)}}]), \; \widehat{\phi_\theta^{(s)}} = \mathbf{U}_s^\intercal \overline{\phi_\theta^{(s)}}, \; \widehat{\psi_\theta^{(s)}} = \mathbf{V}_s^\intercal \overline{\psi_\theta^{(s)}}.
\end{gathered}
\label{normalization_post_training}
\end{equation}
\par\vspace{-1.5ex}\noindent
\noindent After the normalization, we obtain the leading $K$ eigenvalues $\{\widehat{\sigma_K^{(s)}}\}$ and the corresponding basis functions with $\{\widehat{\psi_\theta^{(s)}}, \widehat{\psi_\theta^{(s)}}\}$. Since they form a decomposition of density ratios, with a sufficiently large bandwidth $K$, local-level density ratios can be approximated by 
\begin{equation}
\begin{gathered}
\widehat{\rho_{s, s+1}} = \widehat{\phi_\theta^{(s)}} {}^\intercal \widehat{\mathbf{\Sigma}_s} {}^{\frac{1}{2}}   \widehat{\psi_\theta^{(s)}} \to \rho_{s, s+1}, \;\; 1 \leq s \leq S-1; \;\; \sum_{s=1}^{S-1} \log \widehat{\rho_{s, s+1}} \to \log \overline{\rho}.
\end{gathered}
\end{equation}
This procedure will produce the spectrum, the basis functions, and the approximated density ratio $\widehat{\rho_{s, s+1}}$ at each scale $s$. A thorough pseudocode for this procedure is presented in Algorithm~4.

\subsection{Generate Local Response of Density Ratios}

Here, we further illustrate the procedure for producing figures associated with telescoping density ratios, as shown in Section~5 of the paper. Based on the estimated density ratios, we are able to capture top-down information from the upper layers down to the bottom layers, by propagating density ratios layer by layer. To accomplish this, we establish a sequence of functions ${\varrho_{s}:\mathcal{X}^{(s)} \rightarrow \mathbb{R}}$ through a recursive procedure. Each function in this series is tasked with evaluating global-to-local dependence relationships at respective local scales.

As we apply each convolution layer to the feature maps derived from its preceding layer, we fix $\mathbf{Z}_s(i, j)$ at the lower scale $s$ and identify the corresponding elements at the higher scale $s+1$ that share a mapping relationship with this element $\mathbf{Z}_s(i, j)$. For this purpose, we define four coordinates:
\vspace{-.5ex} 
\begin{equation}
\begin{gathered}
I_s = [\max(0, i-\Delta H_s+1), \min(i, H_{s+1}-1)], J_s = [\max(0, j-\Delta W_s+1), \min(j, W_{s+1}-1)]. 
\end{gathered}
\end{equation}
\par\vspace{-2ex}\noindent
Elements at the higher scale $s+1$ that share a mapping relationship with this element from the lower scale are located within the bounding box defined by these coordinates. Denote the density ratio between receptive fields of $\mathbf{Z}^{(s)}(i', j')$ and $\mathbf{Z}^{(s+1)}(i, j)$ as $\widehat{\rho_{s, s+1}}((i', j'), (i, j))$. Starting with $\varrho_{S} = 1$, we then implement a recursive procedure:
\begin{equation}
\begin{gathered}
\varrho_{s}(i', j') = \sum_{i \in I_s} \sum_{j\in J_s}\varrho_{s+1} (i, j) \widehat{\rho_{s, s+1}}((i', j'), (i, j)), 1 \leq i'\leq H_s, 1 \leq j'\leq W_s.
\end{gathered}
\label{telescoping_post_training}
\end{equation}
For each $1\leq s\leq S-1$, $\varrho_{s}$ is a heatmap with dimensions $W_s \times H_s$. It localizes the pattern with the most significant statistical dependence to the global scale, as viewed from local scales. We provide the pseudocode for this procedure in Algorithm~5.


\begin{algorithm}
\caption{Retrieve eigenspectrum and basis functions}
\begin{algorithmic}[1] 
\STATE Given any input $\mathbf{X}$; $\mathbf{Z}_1 = \mathcal{F}_\theta^{(1)}(\mathbf{X})$
\FOR{$s = 1, \cdots, S-1$}
\STATE $\mathbf{Z}_{s+1} = \mathcal{F}_\theta^{(s+1)}(\mathbf{Z}_{s})$
\STATE SVD: $(\widehat{\mathbf{R}}_\phi^{(s)})^{-\frac{1}{2}} \widehat{\mathbf{P}}_{\phi, \psi}^{(s)} (\widehat{\mathbf{R}}_\psi^{(s)})^{-\frac{1}{2}} = \mathbf{U}_s \widehat{\mathbf{\Sigma}_s} {}^{\frac{1}{2}} \mathbf{V}_s^\intercal $
\STATE $\widehat{\mathbf{\phi}_\theta^{(s)}}(\mathbf{X}) := \mathbf{U}_s^\intercal (\widehat{\mathbf{R}}_\phi^{(s)})^{-\frac{1}{2}} \mathbf{Z}_{s}$
\STATE $\widehat{\mathbf{\psi}_\theta^{(s)}}(\mathbf{X}) := \mathbf{V}_s^\intercal (\widehat{\mathbf{R}}_\psi^{(s)})^{-\frac{1}{2}} \mathbf{Z}_{s+1}$
\STATE $\widehat{\rho_{s, s+1}} = \widehat{\phi_\theta^{(s)}} {}^\intercal \widehat{\mathbf{\Sigma}_s} {}^{\frac{1}{2}}   \widehat{\psi_\theta^{(s)}}$
\STATE Eigenspectrum (dependence measure): $\widehat{\mathbf{\Sigma}_s} = \text{diag}([\widehat{\sigma_1^{(s)}}, \cdots, \widehat{\sigma_K^{(s)}}])$
\STATE Basis functions (features): $\{\widehat{\mathbf{\phi}_\theta^{(s)}}, \widehat{\mathbf{\psi}_\theta^{(s)}}\}$
\STATE Density ratio approximations: $\widehat{\rho_{s, s+1}}$
\ENDFOR
\end{algorithmic}
\label{algorithm4}
\end{algorithm}

\begin{algorithm}
\caption{Generate local response of density ratios}
\begin{algorithmic}[1] 
\STATE Given any input $\mathbf{X}$; $\mathbf{Z}_1 = \mathcal{F}_\theta^{(1)}(\mathbf{X})$
\FOR{$s = 1, \cdots, S-1$}
\STATE $\mathbf{Z}_{s+1} = \mathcal{F}_\theta^{(s+1)}(\mathbf{Z}_{s})$
\STATE Obtain density ratio approximations: $\widehat{\rho_{s, s+1}}:\mathcal{X}_s\times \mathcal{X}_{s+1}\rightarrow \mathbb{R}$.
\ENDFOR
\STATE Initialize $\varrho_{S} = 1$; Initialize each $\rho_{s}$ to be a $(H_s, W_s)$ heatmap. 
\FOR{$s = S-1, \cdots, 1$}
\STATE For every $\mathbf{Z}_{s}(i, j)$, find its mapped counterparts in $\mathbf{Z}_{s+1}$: $\{\mathbf{Z}_{s+1}(i',j'), i'\in I, j' \in J\}$
\STATE $\varrho_{s}(i, j) = \sum_{i \in I} \sum_{j\in J}\varrho_{s+1} (i', j') \widehat{\rho_{s, s+1}}((i, j), (i', j')), 1 \leq i\leq H_s, 1 \leq j\leq W_s$
\ENDFOR
\end{algorithmic}
\label{algorithm5}
\end{algorithm}

\vspace{-10pt}
\section{Additional Experimental Results}

In the main body of the paper, we illustrated the null-space projections caused by adding supervision, and the similarity between supervised and self-supervised scenarios, by visualizing the cross-layer and cross-supervision spectrum. We have also depicted local density ratio responses for nine CIFAR10 samples at layer L2. In addition to these two experiments, we include additional figures as supplementary.


\subsection{Learning Dynamics of Eigenvalues} 
\begin{figure}[H]
  \centering
  \begin{subfigure}{1\textwidth}
    \includegraphics[width=\linewidth]{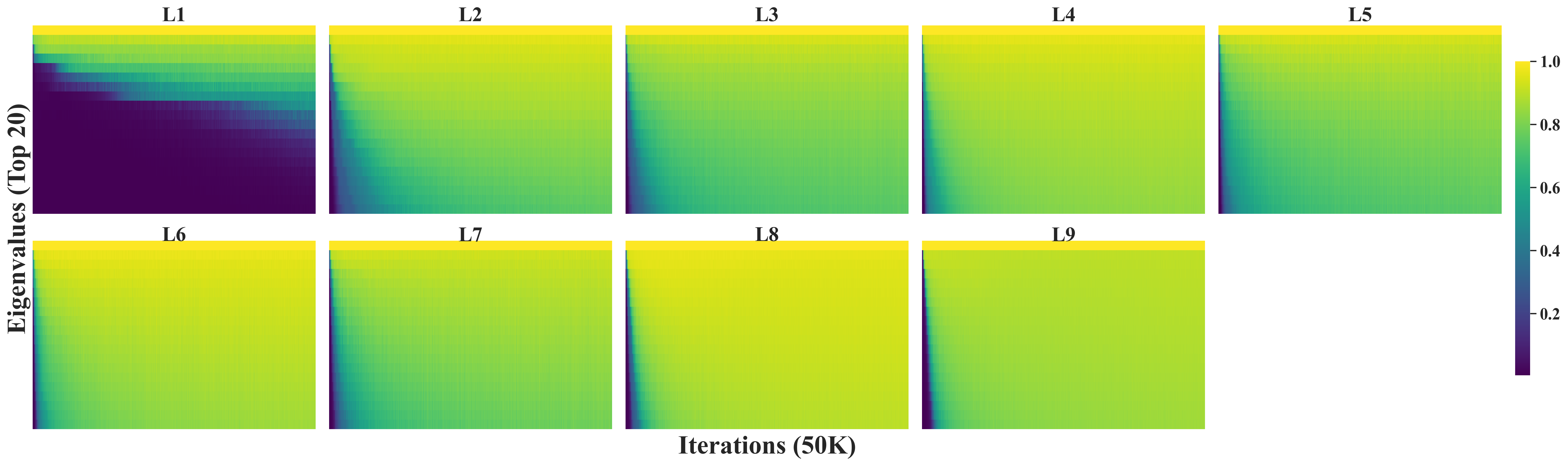}\vspace{-5pt}
    \caption{\small Unsupervised scenario}
  \end{subfigure}\vspace{0pt}
  \begin{subfigure}{1\textwidth}
    \includegraphics[width=\linewidth]{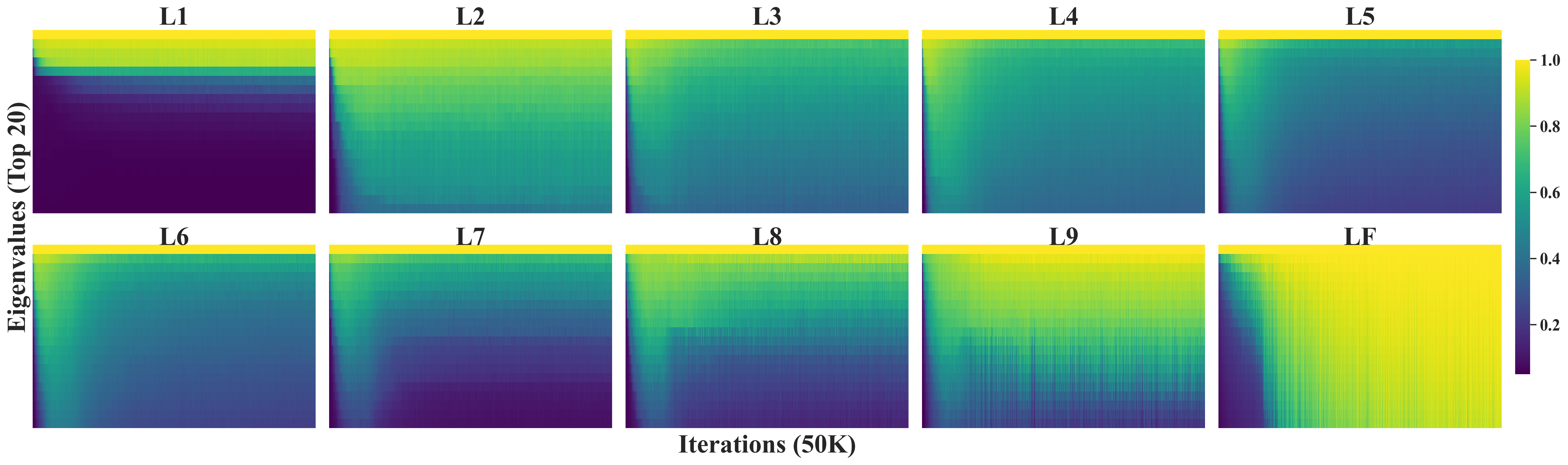}\vspace{-5pt}
    \caption{\small Supervised scenario}
  \end{subfigure}\vspace{0pt}
  \begin{subfigure}{1\textwidth}
    \includegraphics[width=\linewidth]{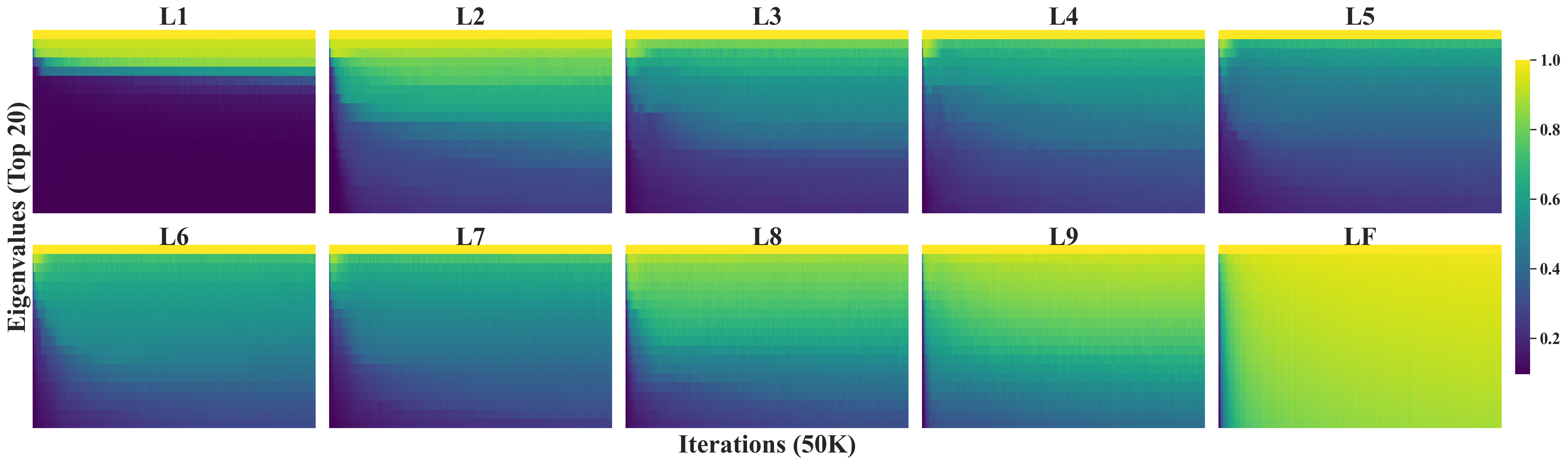}\vspace{-5pt}
    \caption{\small Self-supervised scenario}
  \end{subfigure}\vspace{-5pt}
  \caption{\small Visualizing cross-layer learning dynamics. Each supervision method comprises 10 figures, corresponding to one of 10 layers (from $\mathbf{L1}$ through $\mathbf{L9}$, and $\mathbf{LF}$), except for the unsupervised case without the final layer. The x-axis of each figure represents $50,000$ training iterations, and the y-axis represents the top $20$ eigenvalues. The presented heatmaps illustrate the evolution of these $20$ eigenvalues throughout $50,000$ iterations. The colors in the heatmap correspond to the eigenvalue magnitudes, with lighter areas indicating larger values. In both supervised and self-supervised scenarios, the middle layers display an initial rise in leading eigenvalues, followed by a decrease, suggesting a two-phase learning dynamic: initially capturing the dataset's intrinsic dimensions and then executing null-space projections essential for classification discrimination.}
  \label{learning_dynamics}
\end{figure}

In addition to the spectrum obtained after training the model (corresponding to Fig.~4 and Fig.~5 in the main paper), we also visualize the learning dynamics during training, as seen in Fig.~\ref{learning_dynamics}. In these figures, we represent the learning dynamics of the cross-layer spectrum as heatmaps between all neighboring layers ($\mathbf{L1}, \cdots,  \mathbf{L9}, \mathbf{LF}$) for three types of supervision. The x-axis of each figure represents $50,000$ training iterations, and the y-axis represents the top $20$ eigenvalues. This tracks the evolution of eigenvalues over $50,000$ iterations.

The visualized learning dynamics reveal additional insights. Note that in the heatmap, the color intensity represents the magnitude of eigenvalues. Notably, in both supervised and self-supervised scenarios, the middle layers ($\mathbf{L3}$ through $\mathbf{L8}$) consistently display lighter color at the beginning of training, which indicates there is an initial rise in leading eigenvalues. This increase indicates the stage of capturing the dataset's intrinsic dimensions in the middle layers.

However, as training advances, these leading eigenvalues decrease, suggesting the occurrence of null space projections triggered by the added supervision. Interestingly, this phenomenon is not observed in the unsupervised scenario. This pattern of an initial increase followed by a decrease in dependencies indicates the two phases in the learning dynamics, corresponding to the internal and external sources of dependencies. This behavior is also absent in the initial and final layers since they directly interact with inputs and targets. This effect could be crucial for a better understanding of neural networks.

The figure also highlights the differences between supervised and self-supervised learning. In the middle layers, observe that self-supervised learning only has a marginal decrease after the initial stage. 
In contrast, supervised learning exhibits a more significant and enduring decrease in eigenvalues, indicating a greater influence of null space projections and a diminished ability to capture the intrinsic dimensions of the data, compared to self-supervised learning.

Furthermore, this visualization also uncovers insights beyond the scope of a single-variate dependence measure. Observe that in the supervised scenario, the leading eigenvalues at the final layer are nearly the maximal value $1$, and exhibit more instability with noise. These leading eigenvalues at the final layer appear to propagate to lower layers, suppressing the smaller eigenvalues in the spectrum. This causes the discrimination between small and large eigenvalues in the middle layer, which is necessary for the network to perform the classification task. The spectrum for the self-supervised scenario is much more stable, smooth, and does not exhibit such a discrimination effect between large and small eigenvalues.
\vspace{-5pt}
\subsection{Basis Functions} Another critical component in our proposed decomposition scheme is the eigenfunctions associated with the visualized spectrum, obtained through Algorithm~4.

\begin{figure}[h]
  \centering
  \begin{subfigure}{.3\textwidth}
    \includegraphics[width=\linewidth]{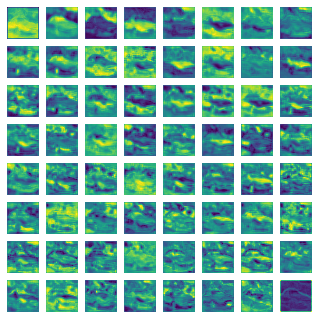}\vspace{-5pt}
    \caption{\small Layer $\mathbf{L2}$}
  \end{subfigure}\hspace{6pt}
  \begin{subfigure}{.3\textwidth}
    \includegraphics[width=\linewidth]{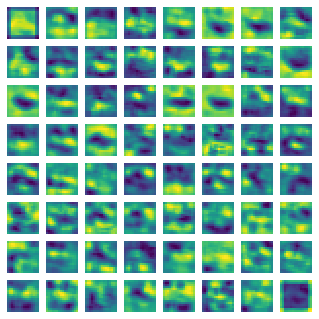}\vspace{-5pt}
    \caption{\small Layer $\mathbf{L4}$}
  \end{subfigure}\hspace{6pt}
  \begin{subfigure}{.3\textwidth}
    \includegraphics[width=\linewidth]{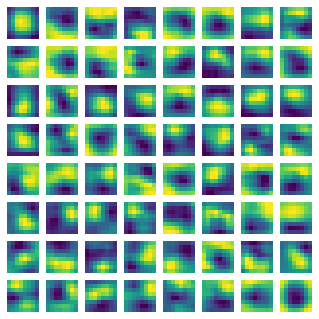}\vspace{-5pt}
    \caption{\small Layer $\mathbf{L6}$}
  \end{subfigure}\vspace{-9pt}
  \caption{\small Visualizing basis functions for one car sample from CIFAR10, at layers $\mathbf{L2}$, $\mathbf{L4}$ and $\mathbf{L6}$, in the unsupervised scenario, each representing a set of orthonormal features at a hierarchical scale. }
  \label{basis_functions}
\end{figure}

By normalizing the feature maps of each layer, each element in the normalized feature maps can be viewed as the corresponding eigenfunction on the respective field in the images, at the particular hierarchical scale. Visualizing these normalized feature maps allows us to observe the evolution of basis functions across the spatial domain. In Fig.~\ref{basis_functions}, we visualize the evolution of the full $64$ eigenfunctions for one selected car sample from CIFAR10 at layer $\mathbf{L2}$, $\mathbf{L4}$ and $\mathbf{L6}$ in the unsupervised scenario.

Fig.~\ref{basis_functions} validates the diversity of the learned basis functions, demonstrating their effectiveness as features at various hierarchical scales. Another observation, which aligns with conventional FMCA, is that the decomposition involving density ratios also adheres to the principles of any orthonormal decomposition. The spatial domain evolution incorporates both high and low frequencies. Generally, basis functions for large eigenvalues tend to represent low-frequency components, and those for small eigenvalues represent high-frequency components.

The features at different hierarchy levels also correspond to the range of dependency relationships. Observations indicate that lower hierarchies capture short-range and higher hierarchies capture long-range relationships. Together, they form a diverse image representation. 

\subsection{Local Density Ratio Responses} In addition to the $9$ samples at layer $\mathbf{L2}$ shown in the main paper, we illustrate local density ratio responses for 20 additional CIFAR10 samples (Fig.~3) across layers $\mathbf{L2}$, $\mathbf{L4}$ and $\mathbf{L6}$ in Fig.~4. Their generation procedure has been thoroughly described in Algorithm~5, which involves constructing density ratio approximations with spectrum (Fig.~1) and eigenfunctions (Fig.~2), and propagating density ratios layer by layer from top to bottom. 

\label{Sec_basis}

\newpage

\vspace{-25pt}%

\begin{SCfigure}[][h]
  \centering
    \includegraphics[width=.3\linewidth]{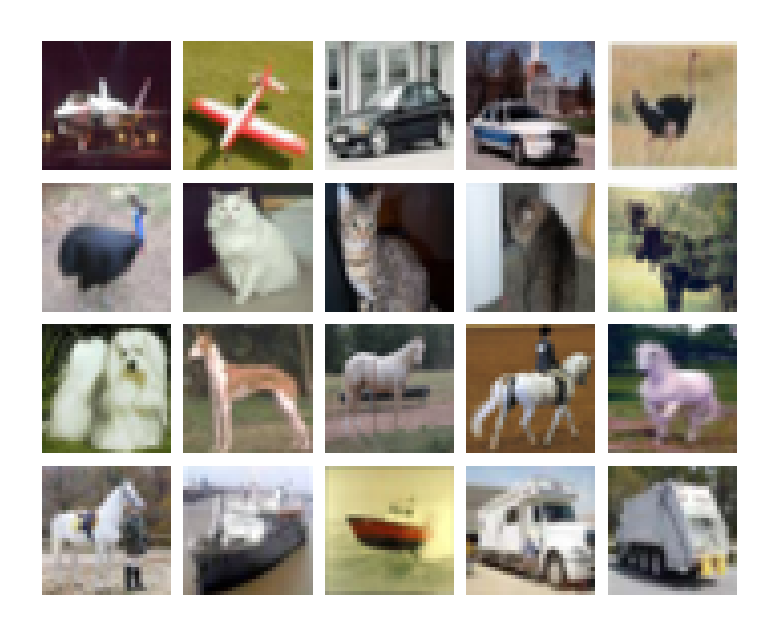}
  \caption{\small The $20$ samples from CIFAR10 we use for comparing local density ratio responses in Fig.~4.}
  \label{original_samples}
\end{SCfigure}
\begin{figure}[H]
\vspace{-20pt}
  \centering

  \begin{subfigure}{.3\textwidth}
    \includegraphics[width=\linewidth]{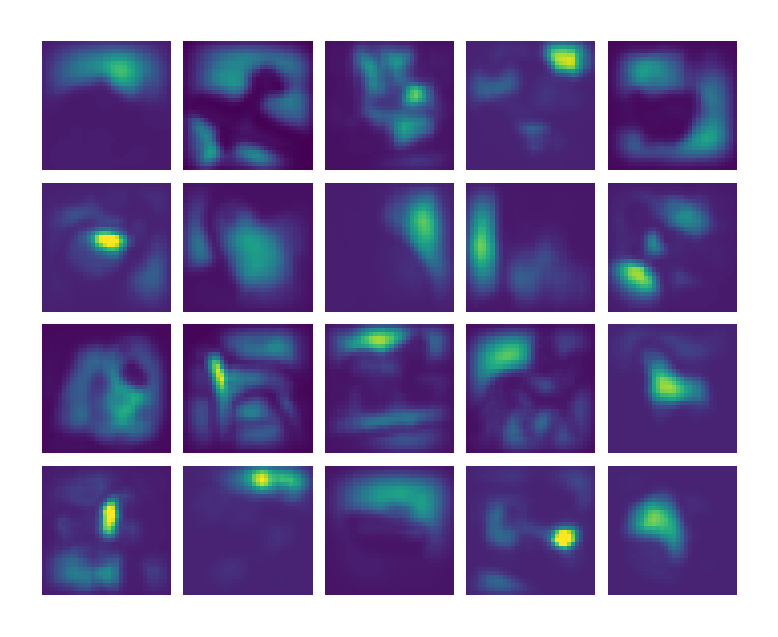}\vspace{-8pt}
    \caption{Unsupervised - $\mathbf{L2}$}
  \end{subfigure}\hspace{0pt}
  \begin{subfigure}{.3\textwidth}
    \includegraphics[width=\linewidth]{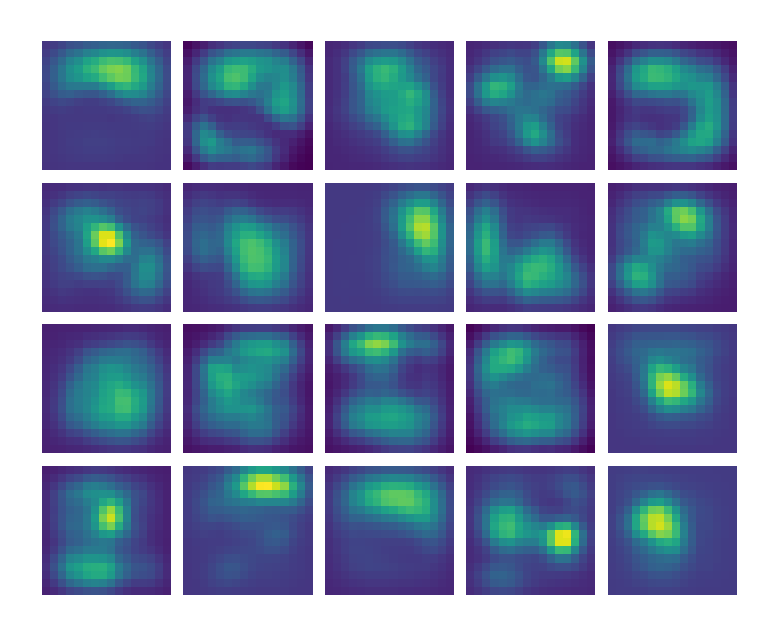}\vspace{-8pt}
    \caption{Unsupervised - $\mathbf{L4}$}
  \end{subfigure}\hspace{0pt}
  \begin{subfigure}{.3\textwidth}
    \includegraphics[width=\linewidth]{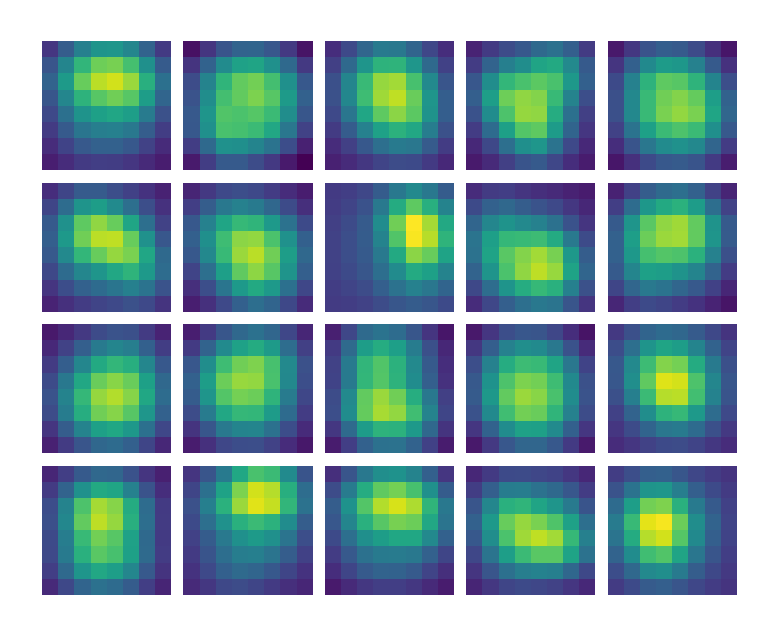}\vspace{-8pt}
    \caption{Unsupervised - $\mathbf{L6}$}
  \end{subfigure}\vspace{-2pt}
  \begin{subfigure}{.3\textwidth}
    \includegraphics[width=\linewidth]{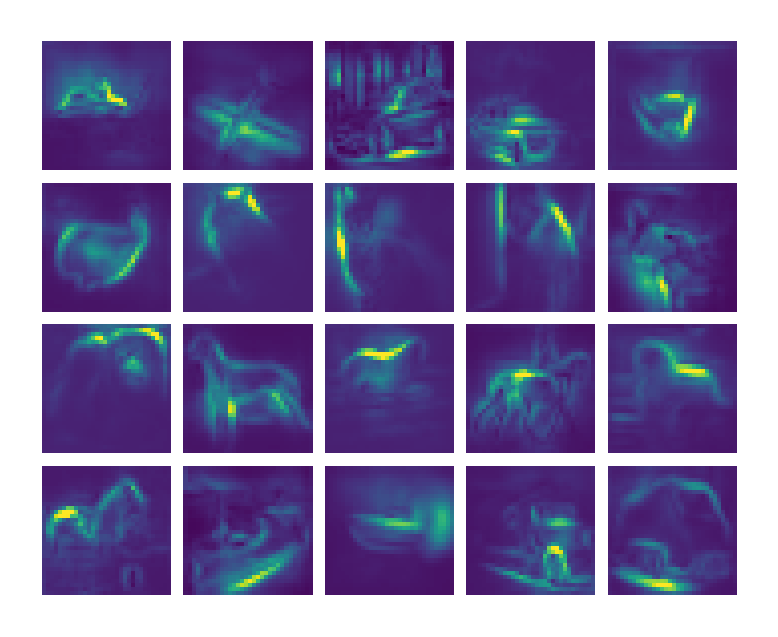}\vspace{-8pt}
    \caption{Supervised - $\mathbf{L2}$}
  \end{subfigure}\hspace{0pt}
  \begin{subfigure}{.3\textwidth}
    \includegraphics[width=\linewidth]{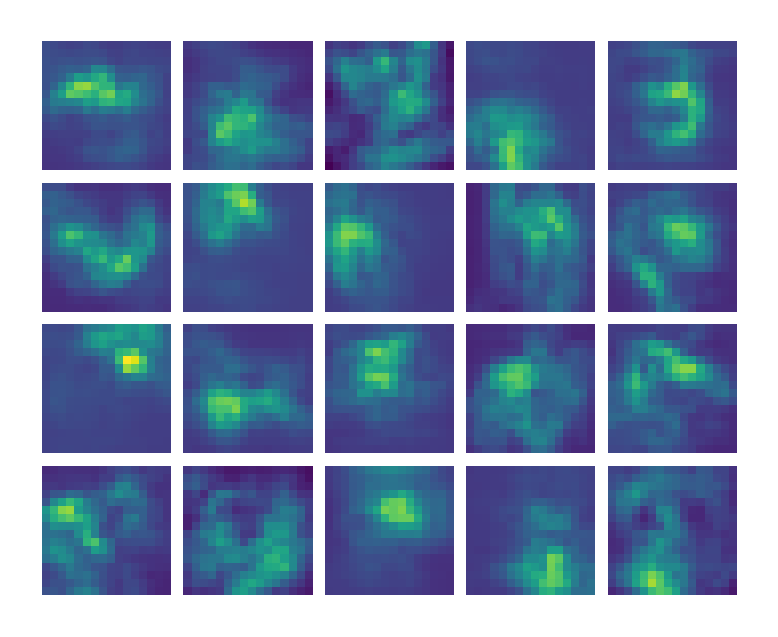}\vspace{-8pt}
    \caption{Supervised - $\mathbf{L4}$}
  \end{subfigure}\hspace{0pt}
  \begin{subfigure}{.3\textwidth}
    \includegraphics[width=\linewidth]{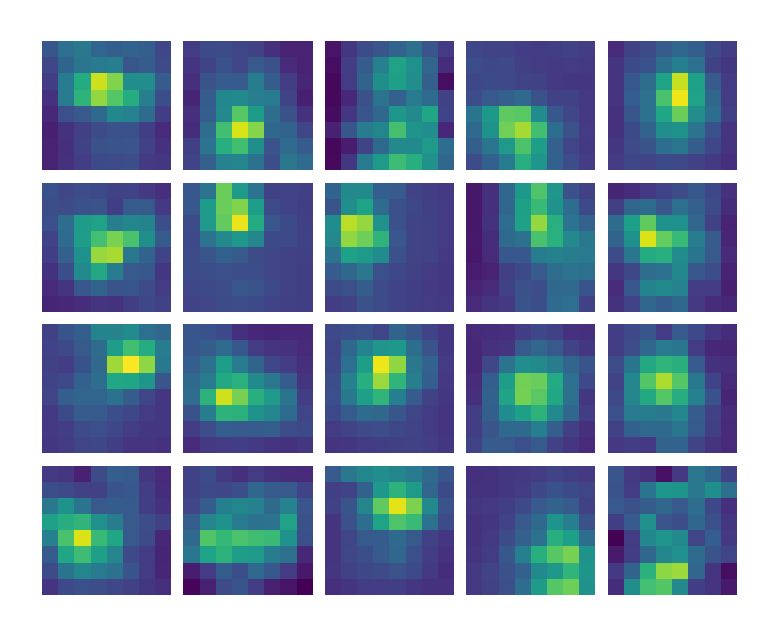}\vspace{-8pt}
    \caption{Supervised - $\mathbf{L6}$}
  \end{subfigure}\vspace{-2pt}

  \begin{subfigure}{.3\textwidth}
    \includegraphics[width=\linewidth]{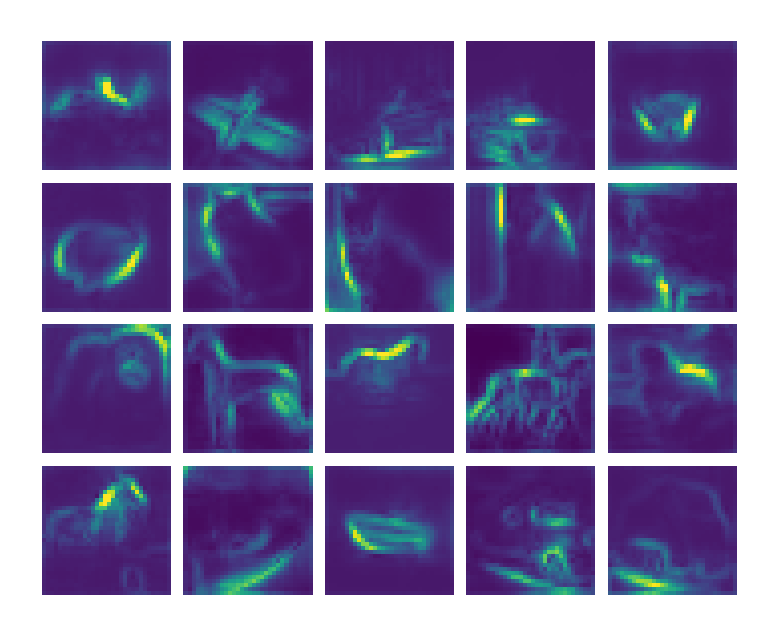}\vspace{-8pt}
    \caption{Self-supervised - $\mathbf{L2}$}
  \end{subfigure}\hspace{0pt}
  \begin{subfigure}{.3\textwidth}
    \includegraphics[width=\linewidth]{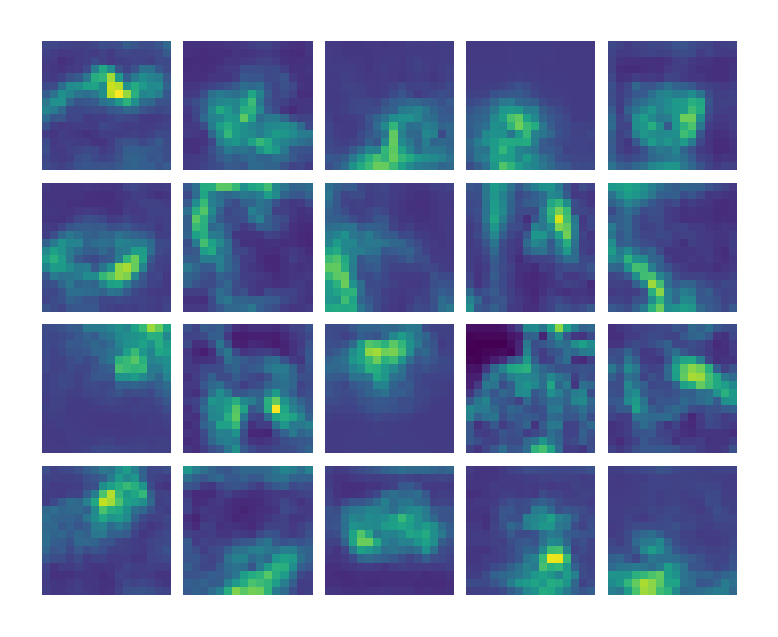}\vspace{-8pt}
    \caption{Self-supervised - $\mathbf{L4}$}
  \end{subfigure}\hspace{0pt}
  \begin{subfigure}{.3\textwidth}
    \includegraphics[width=\linewidth]{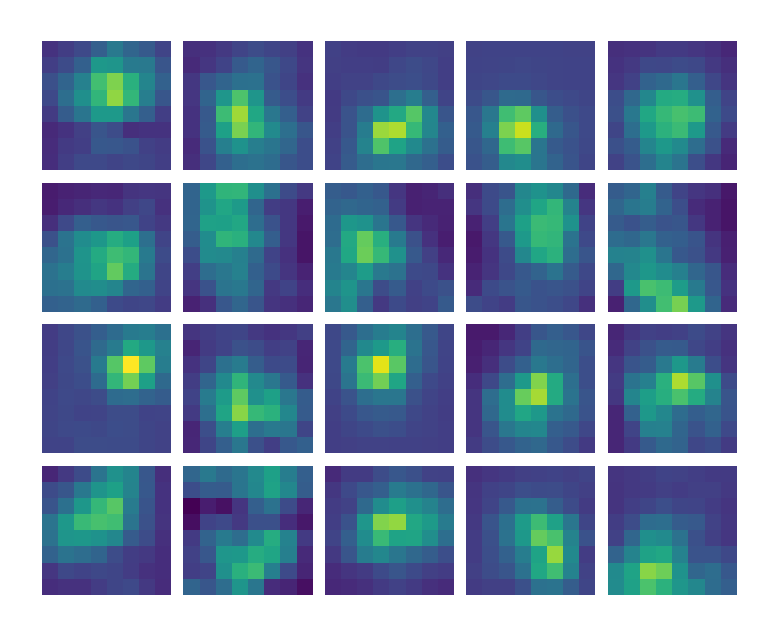}\vspace{-8pt}
    \caption{Self-supervised - $\mathbf{L6}$}
  \end{subfigure}\vspace{-5pt}
  \caption{Local density ratio responses are compared for 20 samples from CIFAR10 at layers $\mathbf{L2}$, $\mathbf{L4}$ and $\mathbf{L6}$, under all three supervision scenarios. The responses reveal consistent patterns across different network layers. The procedure of propagating density ratios effectively allows bottom layers, which are conventionally viewed as having high-resolution but low-level information, to reflect high-level information. This comparison also further confirms the close resemblance between supervised and self-supervised learning, not only at the bottom layers but across all internal layers.}
  \label{activate}
\end{figure}

As discussed in the main paper, adding supervision at the lower layer $\mathbf{L2}$ effectively highlights the most informative regions in the images, such as object boundaries and interactions between different parts of an object. By examining responses across various layers, we can clearly see consistency in the patterns. The key difference is that upper layers tend to represent high-level object information, capturing long-range dependencies, while lower layers focus more on the details. This approach, instead of relying solely on the network's final layer and ignoring all internal outputs, generates a multi-scale representation. Propagating density ratios from upper to lower layers improves the representation by adding details that upper layers, due to their lower resolution, cannot capture.

Inherent in the conventional neural network topology is the property that the bottom layers have high resolution but only contain low-level information, while the top layers have low resolution, despite capturing high-level information. On the contrary, our generated density ratio responses effectively enable the bottom layers to also contain high-level information.

In addition, we present $20$ more car samples visualized at layer $\mathbf{L2}$ for self-supervised learning in Fig.~5. Interestingly, we find that the most frequent pattern for the car class is the boundary of the wheels and the interactions between the car wheels and the ground. Fig.~5 demonstrates that these patterns are indeed consistent within the class. 

\begin{SCfigure}[][h]
  \centering
  \begin{subfigure}{.33\textwidth}
    \includegraphics[width=\linewidth]{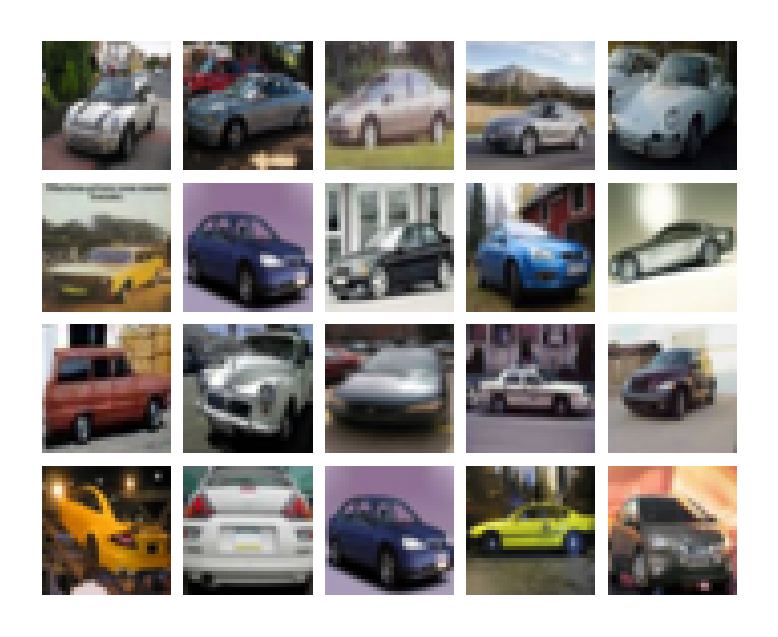}\vspace{-8pt}
    \caption{Randomly Picked $20$ Car Samples}
  \end{subfigure}\hspace{0pt}
  \begin{subfigure}{.33\textwidth}
    \includegraphics[width=\linewidth]{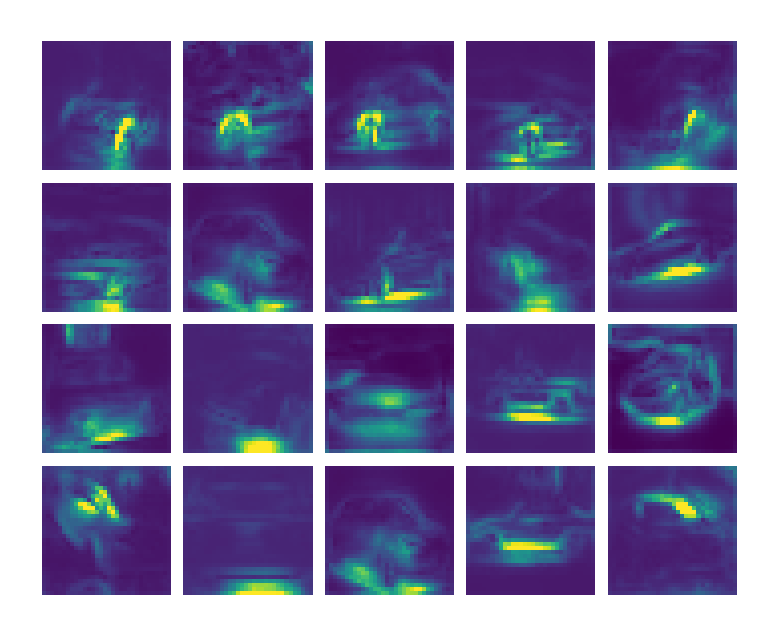}\vspace{-8pt}
    \caption{Self-supervised - $\mathbf{L2}$}
  \end{subfigure}
  \caption{Comparing local density ratio responses for $20$ samples from CIFAR10 car class, at layer $\mathbf{L2}$, in self-supervised scenario. }
  \label{activate_car}
\end{SCfigure}




\section{Implementation Details}

Now we move to the implementation details for the experiments, particularly the used neural network structures and the important hyperparameters for reproducing the experiments. 

\subsection{Network Structure} 

To obtain the optimal performance for self-supervised learning, we adopt the commonly-used ResNet-18 backbone, following the same implementations in~\cite{susmelj2020lightly}. All experiments utilize this ResNet-18 backbone with a consistent batch size of $64$, an SGD optimizer, a learning rate of 0.06, and a momentum of $0.9$. These settings stay consistent with benchmark models.

To obtain the best visualization for analysis purposes, we modify the backbone to a CNN with blocks. The first block $\mathbf{L0}$ (see Table~\ref{blockL0}), employs only $1\times 1$ kernels, transforming input images $\mathbf{X}$ into feature maps $\mathbf{Z}_1$. The blocks $\mathbf{L1}$ through $\mathbf{L8}$ (see Table~\ref{layer_L1_TO_L8}), include two layers with $1\times1$ kernels and one layer with a $3\times3$ kernel in between, making it a universal approximator for $3\times 3$ patches. The block $\mathbf{L9}$ (see Table~\ref{layer_L9}), has an additional average pool at the end, such that the output dimension is $1$. 

The block responsible for constructing the external cost, pertinent to both performance and spectrum analysis, utilizes the network described in Table~\ref{L9}. The input for this block is the concatenation (in the channel dimensions) of feature maps for a group of $9$ samples. For the best performance in self-supervised learning, we enhance the feature dimensions from $64$ to $128$.

Blocks $\mathbf{L0}$ to $\mathbf{LF}$ for spectrum analysis are optimized using an Adam optimizer with a learning rate of $0.0001$, $\beta_1 = 0.5$, $\beta_2 = 0.9$, and a batch size of $32$.

\begin{table}[H]
\centering
\begin{tabular}{@{}lcccccc@{}}
\toprule
\textbf{Layer} & \textbf{In Ch.} & \textbf{Out Ch.} & \textbf{Kernel Size} & \textbf{Padding} & \textbf{Input} & \textbf{Output}\\ \midrule
\rowcolor[HTML]{EFEFEF} 
PaddingLayer & 64 & 64 & - & 1 & $\mathbf{X}$ & -\\ 
NoiseChannel & 64 & 84 & - & - & - & -\\ 
Conv2D & 84 & 200 & 1x1 & 0 &  - & - \\ 
BatchNorm2d & 200 & 200 & - & - & - & - \\ 
ReLU & 200 & 200 & - & - & - & -\\
\rowcolor[HTML]{EFEFEF} 
Conv2D & 200 & 200 & 1x1 & 0 & - & -\\
BatchNorm2d & 200 & 200 & - & - & - & -\\ 
ReLU & 200 & 200 & - & - & - & -\\
\rowcolor[HTML]{EFEFEF} 
Conv2D & 200 & 64 & 1x1 & 0 & - & -\\
BatchNorm2d & 64 & 64 & - & - & - & -\\ 
Sigmoid & 64 & 64 & - & - & - & $\mathbf{Z}_1$\\
\bottomrule
\end{tabular}\vspace{-5pt}
\caption{Blocks $\mathbf{L0}$. All convolution kernels are $1\times 1$. No internal costs. }
\label{blockL0}
\end{table}

\begin{table}[H]
\centering
\begin{tabular}{@{}lcccccc@{}}
\toprule
\textbf{Layer} & \textbf{In Ch.} & \textbf{Out Ch.} & \textbf{Kernel Size} & \textbf{Padding} & \textbf{Input} & \textbf{Output}\\ \midrule
\rowcolor[HTML]{EFEFEF} 
PaddingLayer & 84 & 84 & - & 1 & - & $\mathbf{Z}_{s}$\\ 
NoiseChannel & 64 & 84 & - & - & $\mathbf{Z}_{s}$ & -\\ 
Conv2D & 84 & 200 & 1x1 & 0 & -&-\\ 
BatchNorm2d & 200 & 200 & - & - & -&-\\ 
ReLU & 200 & 200 & - & - & -&-\\
\rowcolor[HTML]{EFEFEF} 
Conv2D & 200 & 200 & 3x3 & 0 & -&-\\
BatchNorm2d & 200 & 200 & - & - & -&-\\ 
ReLU & 200 & 200 & - & - & -&-\\
\rowcolor[HTML]{EFEFEF} 
Conv2D & 200 & 64 & 1x1 & 0 & -&-\\
BatchNorm2d & 64 & 64 & - & - & -&-\\ 
Sigmoid & 64 & 64 & - & - & -& $\mathbf{Z}_{s+1}$ \\
\bottomrule
\end{tabular}\vspace{-5pt}
\caption{Blocks $\mathbf{L1}$ through $\mathbf{L8}$. The second convolution has a kernel of $3 \times 3$. Internal costs between $\mathbf{Z}_{s}$ and $\mathbf{Z}_{s+1}$. Average pooling between every two blocks.}
\label{layer_L1_TO_L8}
\end{table}\vspace{-10pt}
\begin{table}[H]
\centering
\begin{tabular}{@{}lcccccc@{}}
\toprule
\textbf{Layer} & \textbf{In Ch.} & \textbf{Out Ch.} & \textbf{Kernel Size} & \textbf{Padding} & \textbf{Input} & \textbf{Output}\\ \midrule
\rowcolor[HTML]{EFEFEF} 
PaddingLayer & 64 & 64 & - & 1 & $- $ & $\mathbf{Z}_8$\\ 
NoiseChannel & 64 & 84 & - & - & $\mathbf{Z}_8$ & -\\ 
Conv2D & 84 & 200 & 1x1 & 0 &  - & - \\ 
BatchNorm2d & 200 & 200 & - & - & - & - \\ 
ReLU & 200 & 200 & - & - & - & -\\
\rowcolor[HTML]{EFEFEF} 
Conv2D & 200 & 200 & 1x1 & 0 & - & -\\
BatchNorm2d & 200 & 200 & - & - & - & -\\ 
ReLU & 200 & 200 & - & - & - & -\\
\rowcolor[HTML]{EFEFEF} 
Conv2D & 200 & 64 & 1x1 & 0 & - & -\\
BatchNorm2d & 64 & 64 & - & - & - & -\\ 
Sigmoid & 64 & 64 & - & - & - & $ -$\\
AvgPool & 64 & 64 & 4x4 & - & - & $\mathbf{Z}_9$\\
\bottomrule
\end{tabular}\vspace{-5pt}
\caption{Blocks $\mathbf{L9}$. The final layer is an average pooling such that the output dimension is $1$. Internal costs between $\mathbf{Z}_8$ and $\mathbf{Z}_9$.}
\label{layer_L9}
\end{table}\vspace{-10pt}

\begin{table}[H]
\centering
\begin{tabular}{@{}lcccccc@{}}
\toprule
\textbf{Layer} & \textbf{In Ch.} & \textbf{Out Ch.} & \textbf{Kernel Size} & \textbf{Padding} & \textbf{Input} & \textbf{Output}\\ \midrule
\rowcolor[HTML]{EFEFEF} 
NoiseChannel & 64 & 84 & - & - & $\mathbf{Z}_{9, 1}, \mathbf{Z}_{9, 2}, \cdots, \mathbf{Z}_{9, 9}$ & -\\ 
Conv2D & 84 & 200 & 1x1 & 0 &  - & - \\ 
BatchNorm2d & 200 & 200 & - & - & - & - \\ 
ReLU & 200 & 200 & - & - & - & -\\
\rowcolor[HTML]{EFEFEF} 
Conv2D & 200 & 200 & 3x3 & 0 & - & -\\
BatchNorm2d & 200 & 200 & - & - & - & -\\ 
ReLU & 200 & 200 & - & - & - & -\\
\rowcolor[HTML]{EFEFEF} 
Conv2D & 200 & 64 & 1x1 & 0 & - & -\\
BatchNorm2d & 64 & 64 & - & - & - & -\\ 
Sigmoid & 64 & 64 & - & - & - & $\mathbf{Z}_{10}$\\
\bottomrule
\end{tabular}\vspace{-5pt}
\caption{External block $\mathbf{LF}$. Its input is the concatenation of feature maps for a group of $9$ samples. A batch contains multiple such groups. External costs between $\mathbf{Z}_{9, 1}, \cdots, \mathbf{Z}_{9, 9}$ and $\mathbf{Z}_{10}$.}
\label{L9}
\end{table}

Here we address several important considerations:
\vspace{-5pt}
\begin{itemize}[leftmargin=*]
\item Noise Channel: At the beginning of each block, we add an additional $20$ uniform noise channels, matching the size of the input feature maps. This significantly stabilizes the learning dynamics of eigenvalues. \vspace{-1pt}
\item Padding Layer: We assume all $3\times 3$ convolutions are performed without padding. Padding is applied to the image at the start of each block, and the output of the padding layer is treated as $\mathbf{Z}_s$, used to calculate the costs with the output of this block $\mathbf{Z}{s+1}$. The calculations of ACFs and CCFs take the paddings into account. \vspace{-1pt}
\item Average Pooling: The input size is retained within each block. Average pooling with a kernel size of $2$ is applied every two blocks from $\mathbf{L1}$ to $\mathbf{L8}$. Layer $\mathbf{L9}$ contains a $4\times 4$ average pooling operation at the end of the network. We take the output of this pooling operation, which is one-dimensional, as $\mathbf{Z}_9$. \vspace{-1pt}
\end{itemize} \vspace{-3pt}

\subsection{Other Hyperparameters} 


There are two additional hyperparameters that are crucial for our experiments:
\vspace{-5pt}
\begin{itemize}[leftmargin=*]
\item Regularization parameter: Each time we compute the inverse of any ACFs (e.g., gradient estimation in Algorithm 2), similar to the pseudo-inverse, we add a small diagonal matrix, scaled by a regularization parameter, denoted as $\lambda \mathbf{I}$. This ensures the invertibility of the matrices. We found this constant to be important, and it may impact the learned spectrum. To achieve optimal performance in self-supervised learning, we select $\lambda = 0.1$. For spectrum analysis, we choose $\lambda = 0.001$. 
\item Smoothness of adaptive filters (as referred to Algorithm~2): In addition to Adam, we also add adaptive estimators for the ACFs. As the gradient of the log-determinant involves the matrix inverse, we replace the inverse matrices with the estimated ones. Similarly to Adam, we control the smoothness through a parameter $\beta$, as shown in Algorithm~2. To achieve optimal performance in self-supervised learning, we simply set $\beta = 0$. For spectrum analysis, we choose $\beta = 0.5$. \vspace{-1pt}
\end{itemize} \vspace{-3pt}

